\documentclass[journal]{IEEEtran} 

\ifCLASSINFOpdf
  \usepackage[pdftex]{graphicx}
  \graphicspath{{./img/}}
\else
\fi
\usepackage[cmex10]{amsmath}
\usepackage{url}

\newcommand{\Tref}[1]{Table~\ref{#1}}
\newcommand{\Eref}[1]{Eq.~(\ref{#1})}
\newcommand{\Fref}[1]{Fig.~\ref{#1}}

\newcommand{\Sref}[1]{Section~\ref{#1}}
\usepackage{amssymb}
\DeclareMathOperator*{\argmin}{arg\,min}

\usepackage[center]{subfigure}

\usepackage{color}

\newcount\MyDebugFlag

\ifnum \MyDebugFlag=1
\newcommand{\debug}[1]{\textbf{\textcolor{red}{[\textsc{DEBUG:} #1]}}}
\else
\newcommand{\debug}[1]{}
\fi

\newcommand{\bvec}[1]{\mbox{\boldmath $#1$}}

\begin{document}
%
\title{Sketch-based~Manga~Retrieval using~Manga109~Dataset}
%
%
%

\author{Yusuke~Matsui,~\IEEEmembership{Member,~IEEE,}
        Kota~Ito, 
        Yuji~Aramaki, 
        Toshihiko~Yamasaki,~\IEEEmembership{Member,~IEEE,}
        and~Kiyoharu~Aizawa,~\IEEEmembership{Senior~Member,~IEEE,}
\thanks{Y. Matsui, K. Ito, T. Yamasaki, and K. Aizawa are with the Dept. of Information and Communication Eng.,
and Y. Aramaki is with the Graduate School of Interdisciplinary Information Studies, 
The University of Tokyo, Tokyo, Japan.
E-mail: matsui, ito, aramaki, yamasaki, aizawa@hal.t.u-tokyo.ac.jp}
}

%
%

\markboth{Journal of \LaTeX\ Class Files,~Vol.~11, No.~4, December~2012}%
{Shell \MakeLowercase{\textit{et al.}}: Bare Demo of IEEEtran.cls for Journals}
%



\maketitle


\begin{abstract}
	Manga (Japanese comics) are popular worldwide. However, current e-manga archives offer very limited
	search support, including keyword-based search by title or author, or tag-based categorization.
	To make the manga search experience more intuitive, efficient, and enjoyable, we propose 
	a content-based manga retrieval system. First, we propose a manga-specific image-describing framework. 
	It consists of efficient margin labeling, edge orientation histogram feature description, and approximate nearest-neighbor search using
	product quantization.
	Second, we propose a sketch-based interface as a natural way to
	interact with manga content.
	The interface provides sketch-based querying, relevance feedback, and query retouch.
	For evaluation, we built a novel dataset of manga images,
	Manga109, which consists of 109 comic books of 21,142 pages drawn by professional manga artists.
	To the best of our knowledge, Manga109 is currently the biggest dataset of manga images available for research.
	We conducted a comparative study, a localization evaluation, and a large-scale qualitative study.
	From the experiments, we verified that: (1) the retrieval accuracy of the proposed method is higher than those of previous methods;
	(2) the proposed method can localize an object instance with reasonable runtime and accuracy; and (3) sketch querying is useful for manga search. 
\end{abstract}


%
\IEEEpeerreviewmaketitle

\section{Introduction}
\begin{figure}
	\centering
	\includegraphics[width=1.0\linewidth]{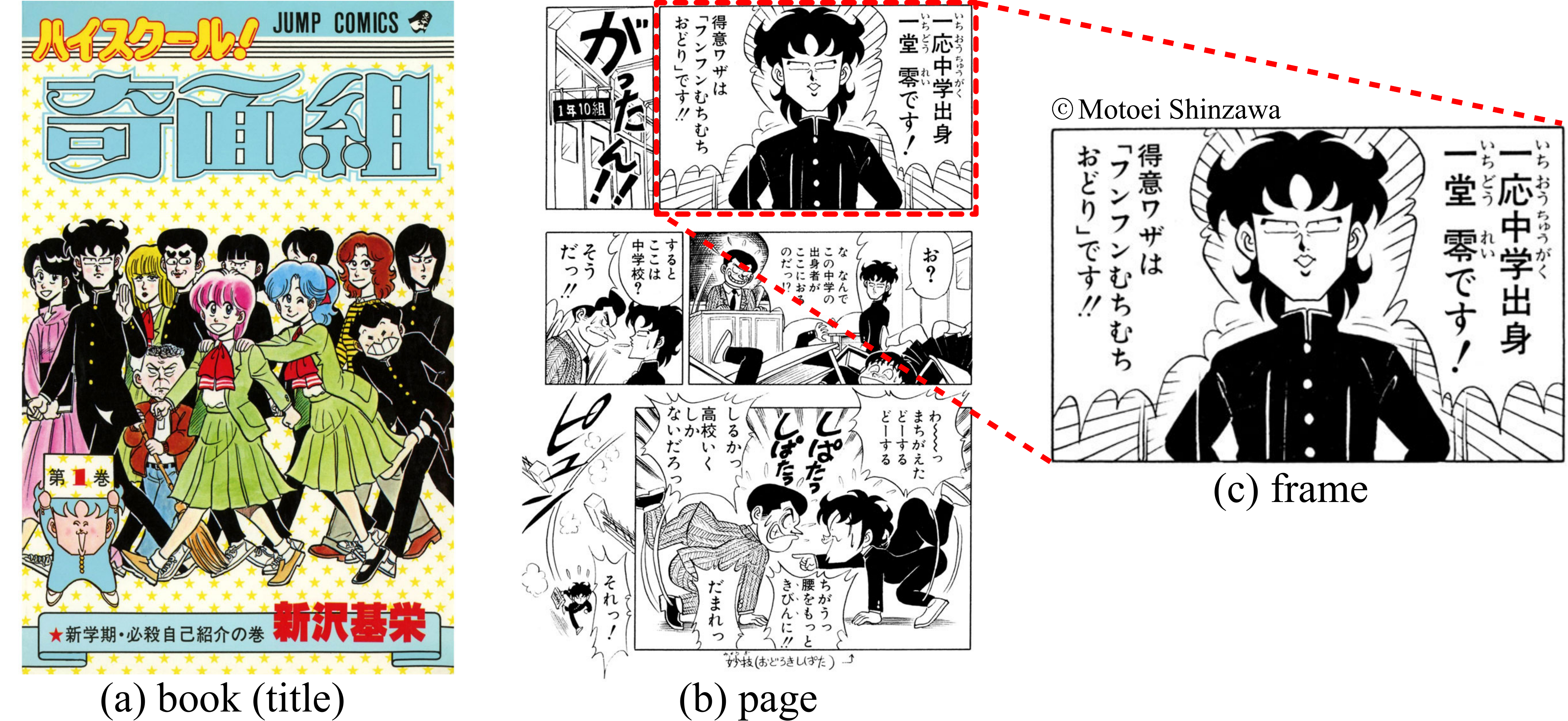}
	\caption{Manga examples. (a) A manga book (title) consists of 200 pages on average. (b) A page (an image) is composed of several frames (rectangular areas). (c) Characters, text balloons, and background are drawn in each frame. Typically, manga are drawn and printed in black and white.}
	\label{fig:teaser}
\end{figure}

\IEEEPARstart{M}{anga} are Japanese black-and-white comics (\Fref{fig:teaser}). They are popular worldwide.
Nowadays, manga are distributed not only in print, but also electronically via online stores such as Amazon Kindle Store\footnote{\url{http://www.amazon.com/Kindle-eBooks/}}.
Some e-manga archives are very large; e.g., the Amazon Kindle store sells more than 130,000 e-manga titles\footnote{Amazon.co.jp Kindle Comic. Retrieved from September 27, 2015, from \url{http://amzn.to/1KD5ZBK}}.
Therefore, users of online stores must retrieve titles from large collections when they purchase new manga.

However, current e-manga archives offer very limited search support, i.e., keyword-based search by title or author, or tag-based categorization. They are not suitable for large-scale search and searches cannot take the images (contents) of manga into consideration. For this reason, applying content-based multimedia retrieval techniques to manga search would have the potential to make the manga-search experience more intuitive, efficient, and enjoyable.

There are three unique challenges associated with content-based manga retrieval.
(1) \textbf{The manga image description problem}.
First, the visual characteristics of manga images are very different from those of naturalistic images. For example, as shown in \Fref{fig:teaser}, manga images are usually line drawings comprising black lines on a flat white background.
Therefore, manga images often do not have varying gradient intensities, unlike naturalistic images.
Traditional local-feature characterizations, such as scale-invariant feature transform (SIFT)~\cite{lowe2004}, may not be suited to describing manga images.
(2) \textbf{Retrieval--localization problem}.
Furthermore, a manga page comprises several frames (rectangular areas). It is therefore necessary to retrieve not only an image, but also a part of the image. This is a combined problem of retrieval and localization, and solutions must allow fast processing to achieve large-scale retrieval.
(3) \textbf{Query modality problem}.
The final challenge is to find a way of efficiently querying manga images.
Suppose that a user wants to retrieve the face of the boy in \Fref{fig:teaser}(c).
Usually, we do not have annotations that describe the specific region,
so we must somehow find a way of querying this region.
Because manga images are mostly black-and-white drawings, users should have access to some form of drawing, such as examples of manga images or sketches of their own. For this reason, it is important to provide a natural sketch-based interface for manga retrieval.

In this paper, we propose a content-based manga retrieval system that addresses the above three challenges.
\begin{itemize}
\item First, we propose a manga-specific image-describing framework, namely an objectness-based edge orientation histogram (EOH) with product quantization. 
The system is composed of three steps: labeling of margin areas, EOH feature description~\cite{levi2004}, and approximate nearest-neighbor (ANN) search using product quantization (PQ)~\cite{jegou2011a}. A comparative study confirmed that the proposed method can efficiently describe manga images, and can yield better accuracy than the state-of-the-art sketch-based retrieval methods~\cite{sun2013a, schneider2014}.
In addition, we evaluated the proposed method in the instance detection framework, and verified that the proposed method can achieve both retrieval and localization from 21,142 manga pages in less than a second with reasonable accuracy.
\item Second, we study the feasibility of a sketch-based interface as a more natural way for people to interact with manga content. Sketch-based interfaces for retrieving multimedia contents have been explored previously~\cite{sun2013a, eitz2011, cao2011}, and we conjecture that they are particularly suited to manga. In addition, based on the sketch interaction, we built two interactive reranking schemes for an intuitive manga search experience: \textbf{relevance feedback} and \textbf{query retouch}.
\end{itemize}

\begin{figure}
	\centering
	\includegraphics[width=1.0\linewidth]{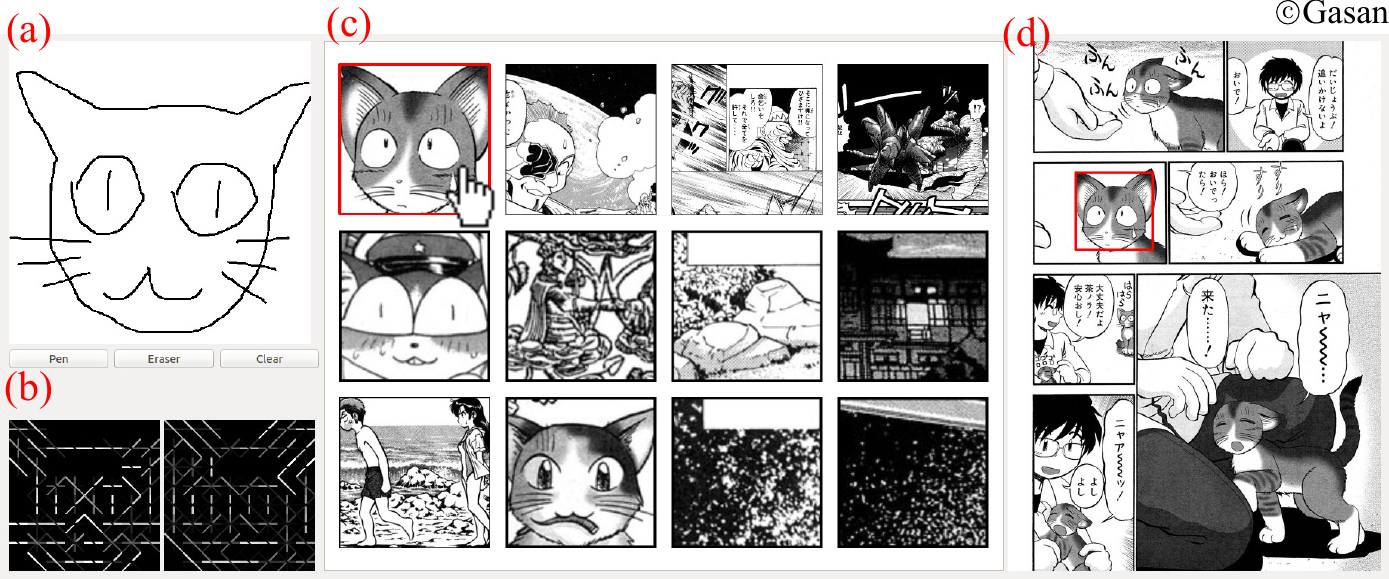}
	\caption{An interface for the retrieval system. (a) A canvas for the user's sketch. (b) Visualized EOH features, where the left figure shows a query and the right shows the retrieved result. (c) Thumbnail windows for the retrieved results. (d) A preview window of a manga page.}
	\label{fig:gui}
\end{figure}

An overview of the GUI implementation of the proposed system is shown in \Fref{fig:gui}. Given a query sketch (\Fref{fig:gui}(a)), the system retrieves similar areas from a manga dataset in real time, and shows the top results in thumbnail windows (\Fref{fig:gui}(c)). The retrieval is performed automatically after each stroke is completed. If a user clicks on one of the retrieved results in the thumbnail windows, a page containing the result is shown in a preview window (\Fref{fig:gui}(d)). In this case, the page containing the first-ranked result is shown. Please see the supplemental video for more details.

For evaluation, we built a novel dataset of manga images drawn by professional manga artists, Manga109, which consists of 109 comic books with a total of 21,142 pages. Manga109 will be very beneficial for the multimedia research community because it has been difficult for researchers to use ``real'' manga in experiments due to the copyright problem. Research so far has been limited to the small scale. We made this dataset with permission from 94 professional creators. The dataset covers a wide range of manga genres, and is publicly available for academic use.

Our contributions are summarized as follows:
\begin{itemize}
	\item We propose a sketch-based manga retrieval system, which enables us to retrieve a manga image in 70 ms from 21,142 pages using a notebook PC. The method consists of a manga-specific image description system, and a sketch-based interactive query scheme, including relevance feedback and query retouch.
	\item We provide a publicly available manga dataset, \textbf{Manga109}, which includes 109 manga titles containing 21,142 pages that will be useful for manga image-processing research.
\end{itemize}

For applications, the proposed method can be integrated in a tablet or a PC to retrieve purchased manga titles stored in each user's computers, or implemented as a Web-based interface to search manga titles at e-manga services such as Amazon. The system is useful in several scenarios. (1) A user wants to find a manga title that contains an object that is hard to represent by words, but easy to represent with a sketch query, such as ``Japanese-style clothing'' (\Fref{fig:relevance_feedback_explain1}). (2) A user does not have a specific title in mind, but tries to find interesting manga, such as one in which cute girls appear. In these cases, our proposed query interactions can be useful (\Fref{fig:relevance_feedback_more1}). (3) After the initial search is performed using text queries, a user can rerank the retrieved results by sketch interactions, e.g., a user selects ``Dragon Ball'' titles by typing ``Dragon Ball'', and then refines results by drawing some characters.

A preliminary version of this work appeared in our recent conference paper~\cite{matsui2014}. This paper contains significant differences: (1) we constructed and released the publicly available manga dataset; (2) we significantly improved the proposed algorithm by leveraging an objectness measure and PQ; and (3) we performed massive experimental evaluations using the new dataset, including comparisons with state-of-the-art methods.

The rest of the paper is organized as follows: \Sref{sec:related} introduces related work. \Sref{sec:manga_retrieval_system} presents our retrieval system and \Sref{sec:query_interaction} shows the proposed query interaction schemes. A novel dataset is introduced in \Sref{sec:db} and evaluations are given in \Sref{sec:exp}.

\section{Related Work}
\label{sec:related}
We discuss related studies in content-based manga retrieval by categorizing them into three challenges: the manga description problem, the retrieval--localization problem, and the querying problem. Additionally, we introduce research related to manga.

\subsection{Manga image description}
In the literature of content-based image retrieval, many methods have been proposed to describe images. A query and dataset images are converted to some form of vector representation to evaluate their similarity. A typical form is the bag-of-features (BoF) representation~\cite{sivic2003}, in which local features such as SIFT~\cite{lowe2004} are extracted from an image and quantized into a sparse histogram. The distance between a query histogram and a dataset histogram can be computed efficiently using an inverted index structure. The BoF form has been studied widely because it achieved successful image description with fast computation.

Various extensions of BoF have been proposed so far, including query expansion~\cite{chum2007}, soft voting~\cite{philbin2008}, and Hamming embedding~\cite{jegou2008}. The state-of-the-art in this line are the vector of locally
aggregated descriptors (VLAD) and Fisher vector (FV) approaches~\cite{jegou2012, spyromitros2014, duan2015}, which are interpreted as generalized versions of BoF~\cite{sanchez2013}. However, all of the above methods were designed for natural images for both query and dataset.
Thus, special attention has been paid to cases in which a query is a line drawing, i.e., a sketch. 

If a query is presented as a form of sketch, several special features tailored for sketch queries have been proposed. BoF using a histogram of oriented gradients~\cite{dalal2005} was proposed~\cite{schneider2014, eitz2011, zhou2012b}. Because sketch images usually have a flat background of uniform white pixels, several methods tried to fill such blank areas with meaningful values and interpreted them as a feature vector; e.g., distance-transformed values from edge pixels~\cite{tseng2012} and Poisson image-editing-based interpolation values~\cite{hu2010, hu2011}. To reduce the impact of noisy edges, a line segment-based descriptor was proposed~\cite{wang2015}. Chamfer matching, which is a successful method for computing the similarity between two contours of objects, was incorporated in the sketch-based retrieval framework~\cite{sun2013a, cao2011, ren2014}.

In the manga feature description task, we compared the proposed method with (1) BoF, (2) large-window FV as the state-of-the-art of BoF-based methods~\cite{schneider2014}, and (3) compact oriented chamfer matching (Compact OCM) as the state-of-the-art of chamfer-based methods~\cite{sun2013a}.


Note that deep-learning techniques are becoming quite popular in image recognition because of their superior performance. These technologies are being applied to the sketch processing. However, the effectiveness of such methods has not yet been discussed sufficiently. For example, Yu and colleagues~\cite{yu2015} reported that traditional methods such as FV are still competitive for a sketch recognition task, and therefore such deep learning-based features are beyond the scope of this paper.

\subsection{Retrieval and localization}
As shown in \Fref{fig:teaser}, our objective is to find not only a similar page, but also identify a similar region in the retrieved image. This kind of task has been tackled in spatial verification-based reranking methods~\cite{lazebnik2006, philbin2007, wu2009, jegou2009, cao2010, zhang2011, zhang2015b}. These methods first find candidate images by BoF, and rerank the top results by spatial information such as relative positions among extracted SIFT features. However, such methods cannot be applied to manga for two reasons. First, the manga image's structure is more complex than that of a natural image. A single manga page usually consists of several frames, each of which is a totally different visual scene (see \Fref{fig:teaser}(b)). Therefore, a manga page can be interpreted as a set of images if we treat each frame as a single image. This is a much harder condition than those usual in image processing tasks such as retrieval or recognition, which assume that at least an image is a single image rather than a set of images.
Second, BoF does not work well for manga as shown in the experimental section.

Another related area is object-detection methods, in which the position of a target object is identified in an input image. A number of methods have been proposed so far. Among them, contour-based object localization methods~\cite{shotton2008, wang2012} are closely related to our manga problem because an object is represented by its contour. In such methods, an instance of a target object such as ``giraffe'' is localized from an image by matching a query contour of giraffe, which is learned from training data. If we apply such methods to all dataset pages and select the best result, we might achieve both retrieval and localization at once. However, this is not realistic because: (1) these methods usually require learning to achieve reasonable performance, i.e., many giraffe images are required to localize an instance of giraffe, but a query is usually a single sketch in our case; and (2) localization usually takes time and cannot be scaled to a large number of images.

The proposed method makes use of a simple approach: window-based matching. It is also slow, but we formulate the search and localization as a single nearest-neighbor problem, and solve it using ANN methods.

\subsection{Querying}
How to query multimedia data has been an important problem in content-based multimedia retrieval~\cite{jiang2013}. Although many approaches have been proposed, including keywords, images, and spatial/image/group predicates~\cite{smeulders2000}, we focus on sketches drawn by users. Sketch-based interaction is the most natural way for humans to describe visual objects, and is widely adapted not only for image retrieval, but also in many applications such as shape modeling~\cite{olsen2009}, image montage~\cite{chen2009}, texture design~\cite{kazi2012}, and as a querying tool for recognizing objects~\cite{schneider2014, eitz2012}.

\subsection{Manga}
From an image-processing perspective, manga has a distinctive visual nature compared with natural images. Several applications for manga images have been proposed: colorization~\cite{qu2006, sykora2009, sato2014}, vectorization~\cite{kopf2012}, layout recognition~\cite{tanaka2007}, layout generation~\cite{hoashi2011, cao2012}, element composition~\cite{cao2014}, manga-like rendering~\cite{qu2008}, speech balloon detection~\cite{rigaud2013}, segmentation~\cite{aramaki2014}, face tailored features~\cite{chu2014}, screen tone separation~\cite{ito2015}, retargeting~\cite{matsui2011}, and manga-like summarization for video~\cite{herranz2012, chu2015, jing2015}. The research group at Universit\'e de La Rochelle constructed a comic database~\cite{guerin2013}, which we mention in \Sref{sec:db}, and analyzed the structure of visual elements in a page to understand the comics content semantically~\cite{rigaud2015}. 

Several studies have explored manga retrieval~\cite{sun2013b, le2015, delalandre2014, weihan2013}. In these methods, inputs are usually not pages but cropped frames or characters, and evaluations were conducted using a small number of examples. In addition, the runtimes of these methods have not been discussed fully. Therefore, it is not known whether these methods can scale to a large dataset. On the other hand, the proposed method can perform a retrieval from 21,142 pages in 70 ms.

\section{Manga Retrieval System}
\label{sec:manga_retrieval_system}

\begin{figure*}
	\centering
	\includegraphics[width=0.8\linewidth]{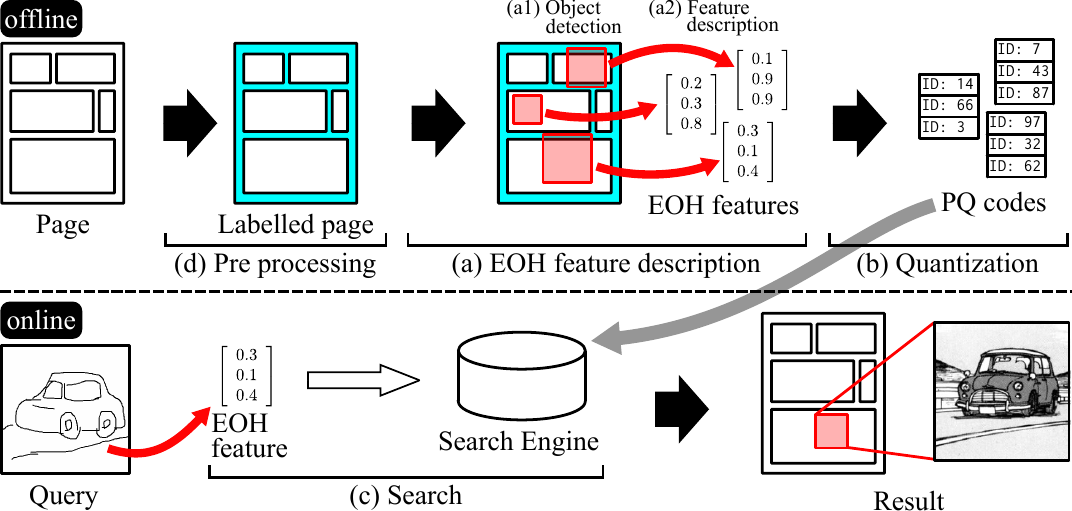}
	\caption{Framework of the proposed system.}
	\label{fig:framework}
\end{figure*}

In this section, we present the proposed manga retrieval system. \Fref{fig:framework} shows the framework of the system.
First, input pages are preprocessed offline (\Fref{fig:framework}(d). This is discussed in \Sref{sec:margin}).
Next, region of interest (ROI) areas are detected and EOH features are extracted (\Fref{fig:framework}(a), as described in \Sref{sec:eoh}).
These features are then compressed into PQ codes
(\Fref{fig:framework}(b), as discussed in \Sref{sec:pq}).
At query time, given a query sketch, similar EOF regions are
retrieved
(\Fref{fig:framework}(c), as discussed in \Sref{sec:search}).
The system is designed to solve both the manga image description and retrieval--localization problems.

The proposed framework is based on the window search paradigm; i.e., several features are extracted from dataset images using a generic object detector, and a query feature is compared with all features.
Compared with the traditional image retrieval problem, such a window system is usually too slow to handle a large number of features.
We formulate the problem as a single ANN search, and solve it using PQ, with effective preprocessing including the labeling of margin areas. Note that we delete screen tone (texture) areas in advance during the preprocessing~\cite{ito2015}.

\subsection{EOH feature description with object windows}
\label{sec:eoh}
\begin{figure}
	\centering
	\subfigure[]{\label{fig:histexample_img} \includegraphics[height=0.3\linewidth]{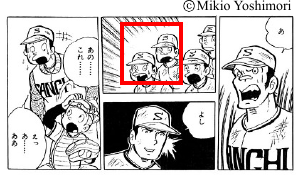}}
	\subfigure[]{\label{fig:histexample_hist} \includegraphics[height=0.3\linewidth]{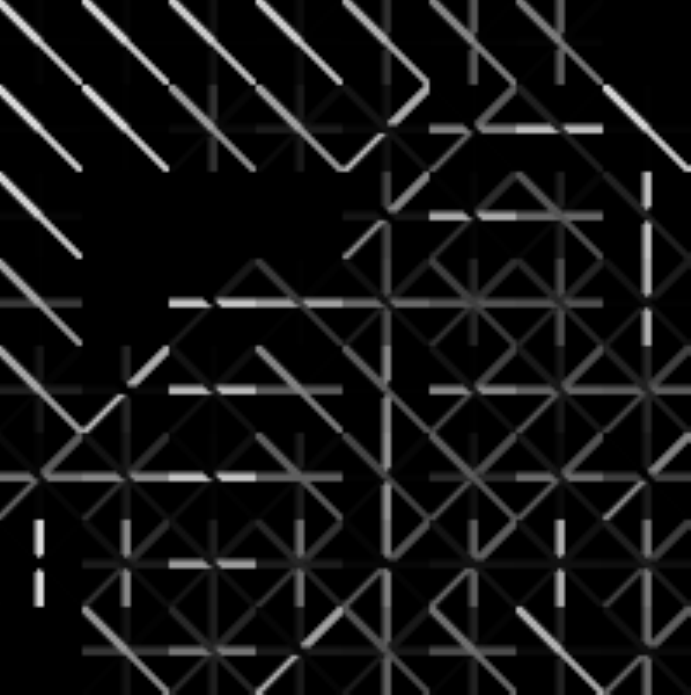}}
	\caption{An example of an EOH feature. (a) An image and a selected area. (b) The visualized EOH features extracted from the area in (a). As can be seen, the visual characteristics of a manga page can be captured using such a simple histogram.}
	\label{fig:histexample}
\end{figure}

We first show a core part of the system, feature description (\Fref{fig:framework}(a)).
We employ EOH features~\cite{levi2004}, which compute an EOH for a square target ROI.
Given a page, candidate areas of objects are detected automatically (\Fref{fig:framework}(a1)),
then EOH features are described from selected areas (\Fref{fig:framework}(a2)).
The retrieval is performed by matching an EOH feature of the query against the EOH features in the dataset.
\Fref{fig:histexample} shows an example of the feature representation. The EOF feature of the red square area in \Fref{fig:histexample}(a) is visualized as shown in \Fref{fig:histexample}(b).

The candidate areas of objects are detected by generic object detectors~\cite{alexe2012, uijlings2013, cheng2014a}, which are successful ways to replace a sliding window paradigm. (\Fref{fig:framework}(a1)).
Intuitively, these detectors compute bounding boxes, which are more likely to contain any kinds of objects. These bounding boxes are considered as candidates for object instances, and a processing task such as image recognition is applied only to the bounding box areas, which greatly reduces computational costs compared with a simple sliding window approach. 

\begin{figure}
	\centering
	\includegraphics[width=1.0\linewidth]{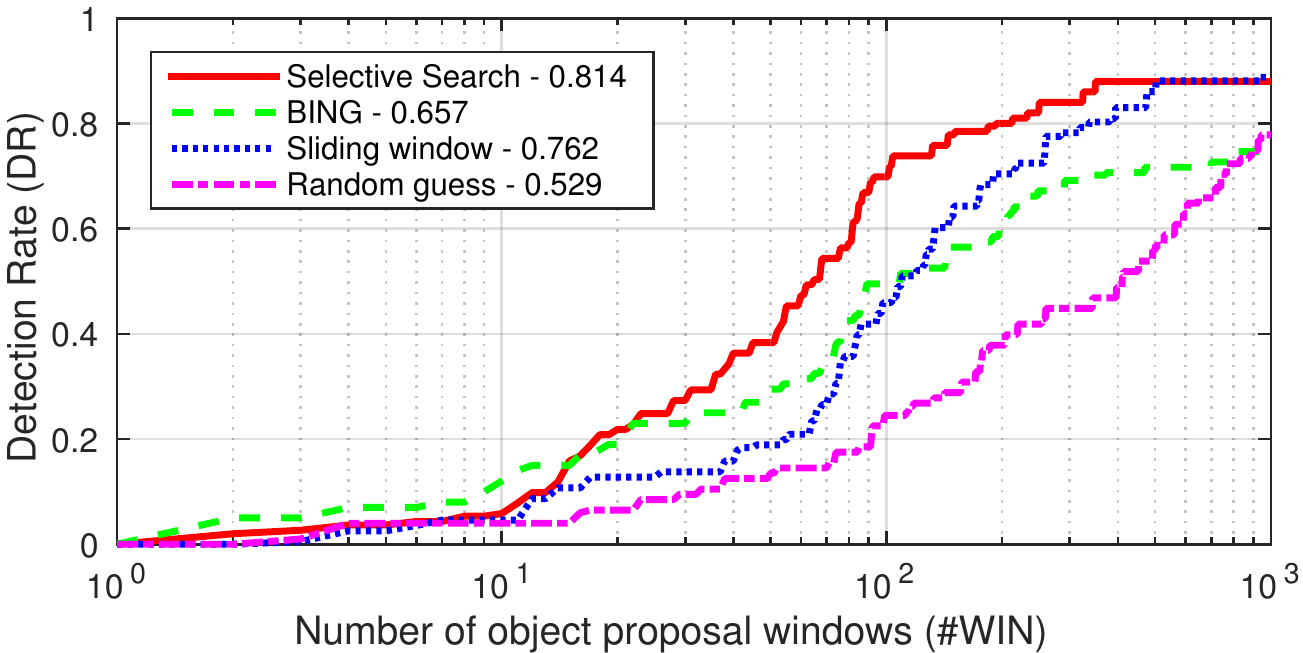}
	\caption{Detection rate (DR) given \#WIN proposals, which is a standard evaluation metric for evaluating objectness~\cite{cheng2014a, alexe2012, uijlings2013}. DR shows the percentage of ground-truth objects covered by proposals, which are generated by each method. An object is considered covered only when the PASCAL overlap criteria is satisfied (intersection over union is more than 0.5~\cite{everingham2010}). The area under the curve (AUC), which is a single-value summary of performance for each curve, is shown on the legends. We evaluated methods using the dataset used for the localization task, see details on \Sref{sec:exp_local}.}
	\label{fig:objectness}
\end{figure}

We found that objectness measures can detect candidate manga image objectives more effectively than the sliding window approach. We constructed an experiment using a small dataset in which ground-truth objects (a head of a man) are annotated. The detection rate of a few methods were evaluated as shown in \Fref{fig:objectness}. BING~\cite{cheng2014a}, selective search~\cite{uijlings2013}, and sliding window (baseline)~\cite{matsui2014} are compared. From the results, we concluded that selective search is the best approach for this dataset. In the rest of this paper, we use selective search for our object detector. Note that a horizontally or vertically long bounding box is divided into several square boxes with overlaps because we restrict the shape of object areas to a square.

After object areas have been detected (typically, 600 to 1000 areas are selected from a page), their EOF features are described (\Fref{fig:framework}(a2)).
The selected square area is divided into $c\times c$ cells. The edges in each cell are quantized into four orientation bins and normalized, and the whole vector is then renormalized. Therefore, the dimension of the EOH vector is $4c^2$.
Note that we discard features in which all elements are zero. For manga, the features should be robust against scale changes because we want to find areas of any size that are similar to the input sketch. For example, if a certain character's face is the target, it can be either small or large. To achieve matching across different scales, we simply accept any sizes of patches, but restrict the shape of an area to square.
Whatever the patch sizes, the same-dimensional EOH features are extracted and used for matching.

By describing EOF features of candidate areas, a page is represented by a set of EOH features:
\begin{equation}
\mathcal{P} = \left \{ \bvec{x}_i \middle | i \in \{1, \dots, N \} \right \} 
\end{equation}
where $\mathcal{P}$ means the page, $\bvec{x}_i$ denotes an EOH feature for the $i$th window, and $N$ is the number of features extracted from the page. Note that integral images are utilized to enable the fast computing of features in the same manner as in \cite{levi2004}. In the comparative study, we show this simple representation achieves better accuracy than previous sketch-based retrieval methods, and we confirm that EOH-based description gives a good solution to the manga image representation problem.

In summary, given a page, candidate areas are detected by selective search, and EOH features are extracted, then the page is represented as a set of EOH features. This window-based representation is simple and efficient at finding a matching area in the page because a feature implicitly contains its position in the page, which is particularly useful for manga pages, and so this approach effectively solves the retrieval--localization problem.

\subsection{Feature compression by product quantization}
\label{sec:pq}
Given a set of EOH features, we compress them into binary codes using PQ~\cite{jegou2011a}, which is a successful ANN method, as shown in \Fref{fig:framework}(b).
Applying PQ greatly reduces the computational cost of matching and 
reduces memory usage.
After the features are compressed to binary codes, the approximate distances between a query vector and dataset codes can be computed efficiently by simple lookup operations (asymmetric distance computation (ADC)~\cite{jegou2011a}). Therefore, the search can be performed efficiently even for large number of dataset vectors.

We now briefly describe the PQ algorithm. 
We represent $\bvec{x}\in\mathbb{R}^D$ as a concatenation of $M$ subvectors: $\bvec{x}~=~[\bvec{x}^1,\dots,\bvec{x}^M]$, where each $\bvec{x}^m \in \mathbb{R}^{D/M}$.
We define a subquantizer $\bvec{q}^m(\bvec{x}^m)$ for $\bvec{x}^m$ as follows:
\begin{equation}
\bvec{x}^m \mapsto
\bvec{q}^m \left ( \bvec{x}^m \right ) =\argmin_{\bvec{c}^m \in \mathcal{C}^m} \left \Vert \bvec{x}^m - \bvec{c}^m \right \Vert.
\end{equation}
The subcodebook $\mathcal{C}^m$ consists of $K$ $D/M$-dimensional centroids learned by $k$-means~\cite{lloyd1982} in advance.
Therefore, $\bvec{q}^m(\bvec{x}^m)$ maps a subvector $\bvec{x}^m$ to a code word $\bvec{c}^m \in \mathcal{C}^m$.

A product quantizer $\bvec{q}(\bvec{x})$ is defined as a concatenation of subquantizers:
\begin{equation}
\bvec{x} \mapsto
\bvec{q} \left ( \bvec{x} \right) =
\left [ \bvec{q}^1 \left ( \bvec{x}^1 \right ), \dots,
\bvec{q}^M \left ( \bvec{x}^M \right ) \right ].
\end{equation}
Therefore, an input $\bvec{x}$ is mapped by $\bvec{q}(\bvec{x})$ to a code word $\bvec{c} = [\bvec{c}^1, \dots, \bvec{c}^M] \in \mathcal{C} = \mathcal{C}^1 \times \dots \times \mathcal{C}^M$.

The resultant code word $\bvec{c}$ is encoded by a tuple $(i^1, \dots, i^M)$ of $M$ subcentroid indices. Because each $i^m$ is an integer, i.e., $i^m \in \{0, \dots, K-1\}$, the code $\bvec{c}$ can be represented by $M\log_2 K$ bits. Typically, $K$ is made a power of 2 so that $\log_2 K$ is an integer. As in other methods, we set $K$ as 256 in this paper ($\bvec{c}^m$ is represented by 8 bit). Note that $M$, which shows the level of quantization, is set as 8 or 16, which yields the length of the code as 64 and 128 bits, respectively.

We now return to our EOH feature description. Recall the page is represented as $\mathcal{P}=\{ \bvec{x}_i| i \in \{1, \dots, N \} \}$. Each feature $\bvec{x}_i$ is quantized as $\bvec{q}(\bvec{x}_i)$, and the page is represented as
\begin{equation}
\mathcal{P}_{PQ} =\left \{ \bvec{q}(\bvec{x}_i) \middle | i \in \{1, \dots, N \} \right \},
\label{eq:p_pq}
\end{equation}
where $\mathcal{P}_{PQ}$ is a set of quantized EOH features for a page. Because each $\bvec{q}(\bvec{x}_i)$ is recorded as a tuple of $M$ subcentroid indices, which take $8M$ bits, $\mathcal{P}_{PQ}$ is stored by $8MN$ bits.

\subsection{Search}
\label{sec:search}
Let us describe how to retrieve the nearest PQ-encoded feature from many manga pages (\Fref{fig:framework}(c)). Suppose there are $P$ manga pages, and the quantized EOH features of the $p$th page are represented as $\mathcal{P}_{PQ}^p =\left \{ \bvec{q}(\bvec{x}_i^p) \middle | i \in \{1, \dots, N^p \} \right \}$, where $N^p$ means the number of EOH features from the $p$th page, and $p \in \{ 1, \dots, P \}$. Given an EOH feature $\bvec{y}$ from a query sketch, the search engine computes the nearest neighbors using:
\begin{equation}
\left < p^*, i^* \right > = \argmin_{ p \in \{1, \dots, P \}, ~i \in \{ 1, \dots, N^p \} } d_{AD} \left ( \bvec{y}, \bvec{x}_i^p \right ),
\end{equation}
where $d_{AD}(\cdot, \cdot)$ measures an approximate Euclidean distance between an uncompressed vector and a compressed PQ code~\cite{jegou2011a}. This can be computed efficiently by lookup operations (ADC~\cite{jegou2011a}).

By computing $d_{AD}$ from $\bvec{y}$ to each $\bvec{x}_i^p$ and selecting the smallest one, $p^*$ and $i^*$ are computed.
This is a simple linear search problem for PQ codes, which can be solved efficiently; e.g., 16 ms for searching one million 64-bit codes.
Then we can say that the $i^*$th feature from the $p^*$th page is the nearest to the query sketch. The point here is that the problem is separated into a feature description and an ANN search. Therefore, even if the number of features is very large, we can employ the ANN techniques. In our system, searching from 21,142 pages takes 793 ms (without parallel implementation) or 70 ms (with parallel implementation) on a single computer using a simple PQ linear scan, which is fast enough. If much faster searches are required, we can employ more sophisticated ANN data structures~\cite{kalantidis2014, babenko2014c}.

\subsection{Skipping margins}
\label{sec:margin}

\begin{figure}
	\centering
	\subfigure[]{\label{fig:framesep_original}
		\includegraphics[width=0.226\linewidth]{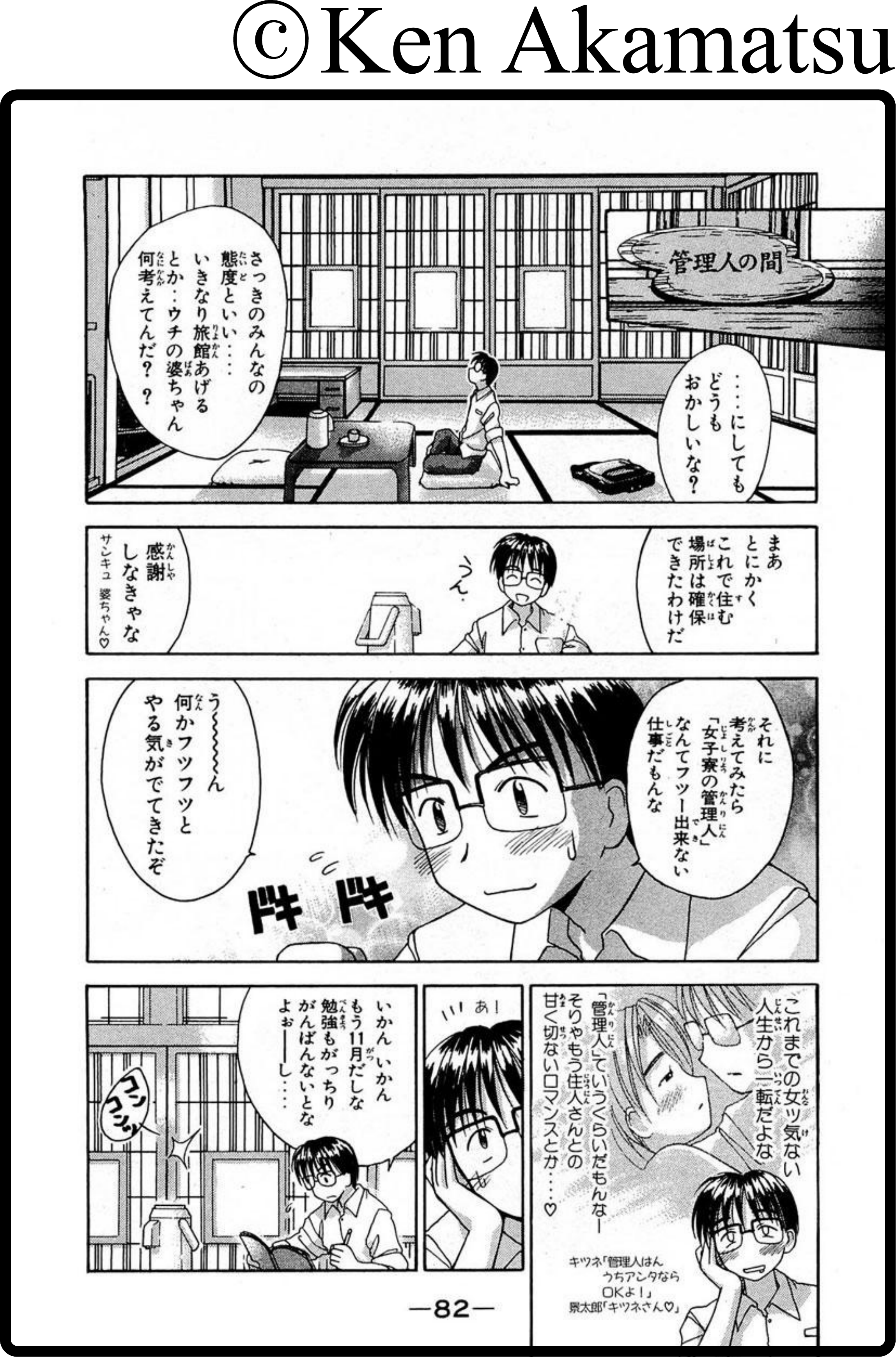}}
	\subfigure[]{\label{fig:framesep_eroded} \includegraphics[width=0.226\linewidth]{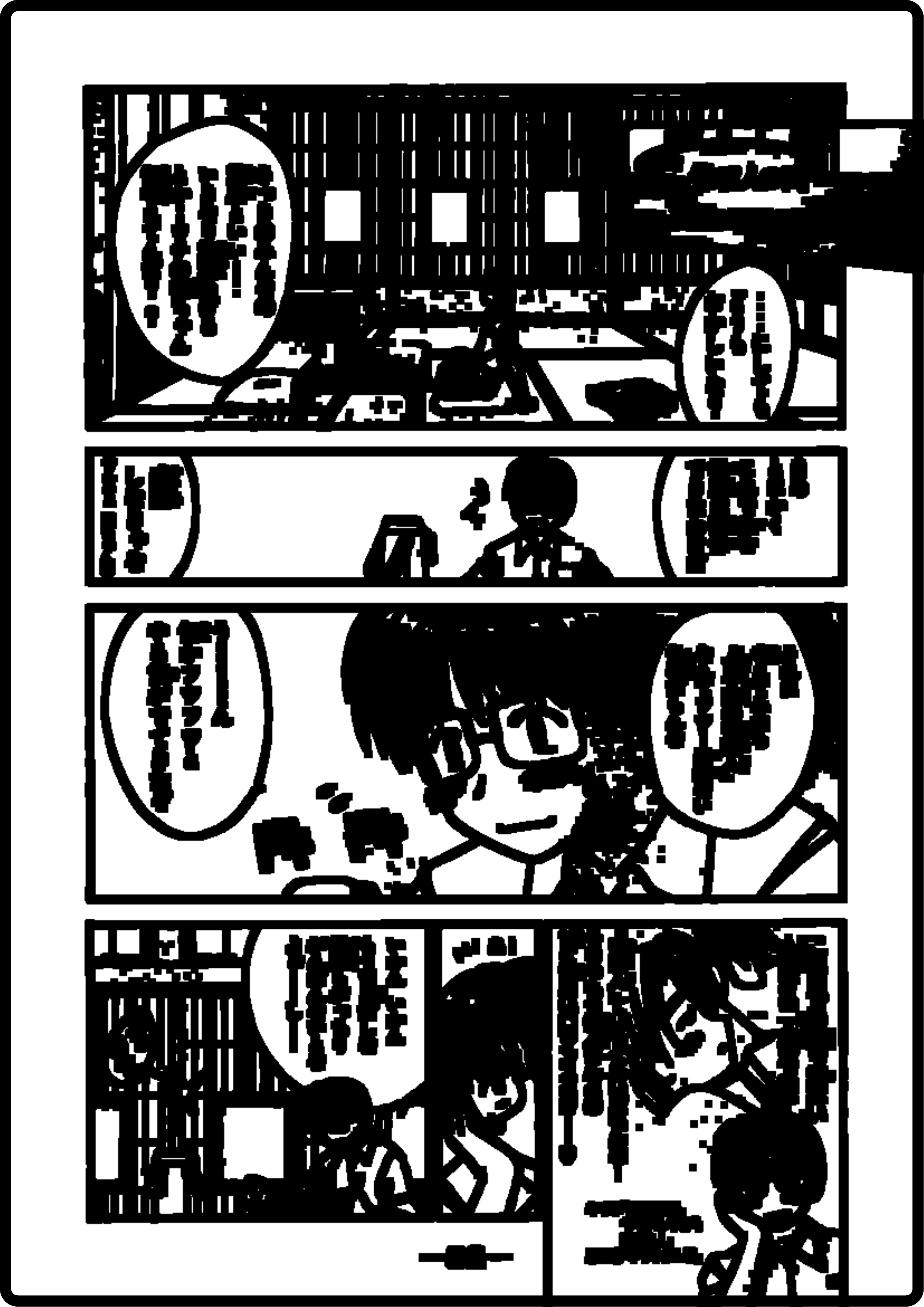}}
	\subfigure[]{\label{fig:framesep_labeled_colored} \includegraphics[width=0.226\linewidth]{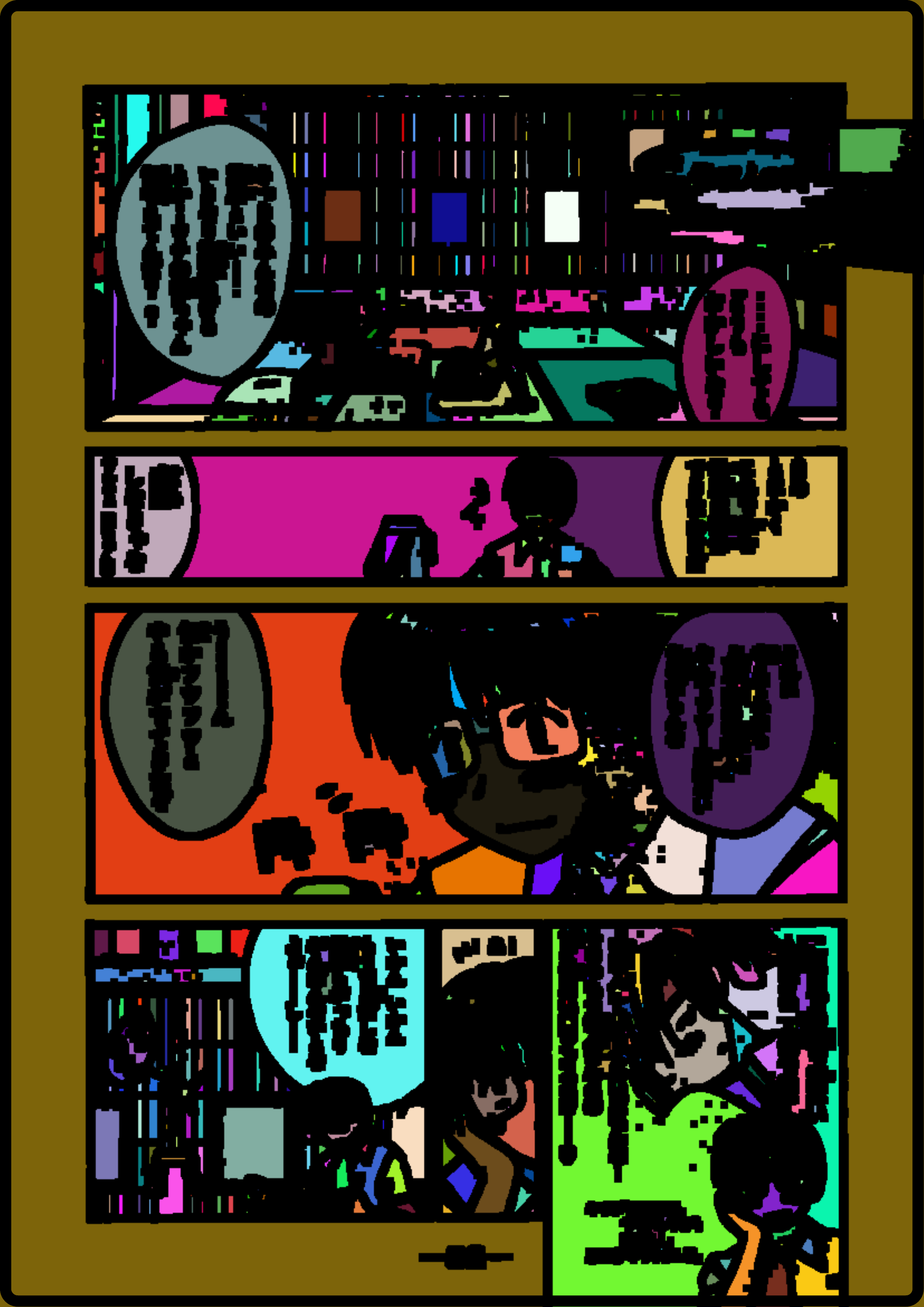}}
	\subfigure[]{\label{fig:framesep_labeled_one_colored} \includegraphics[width=0.226\linewidth]{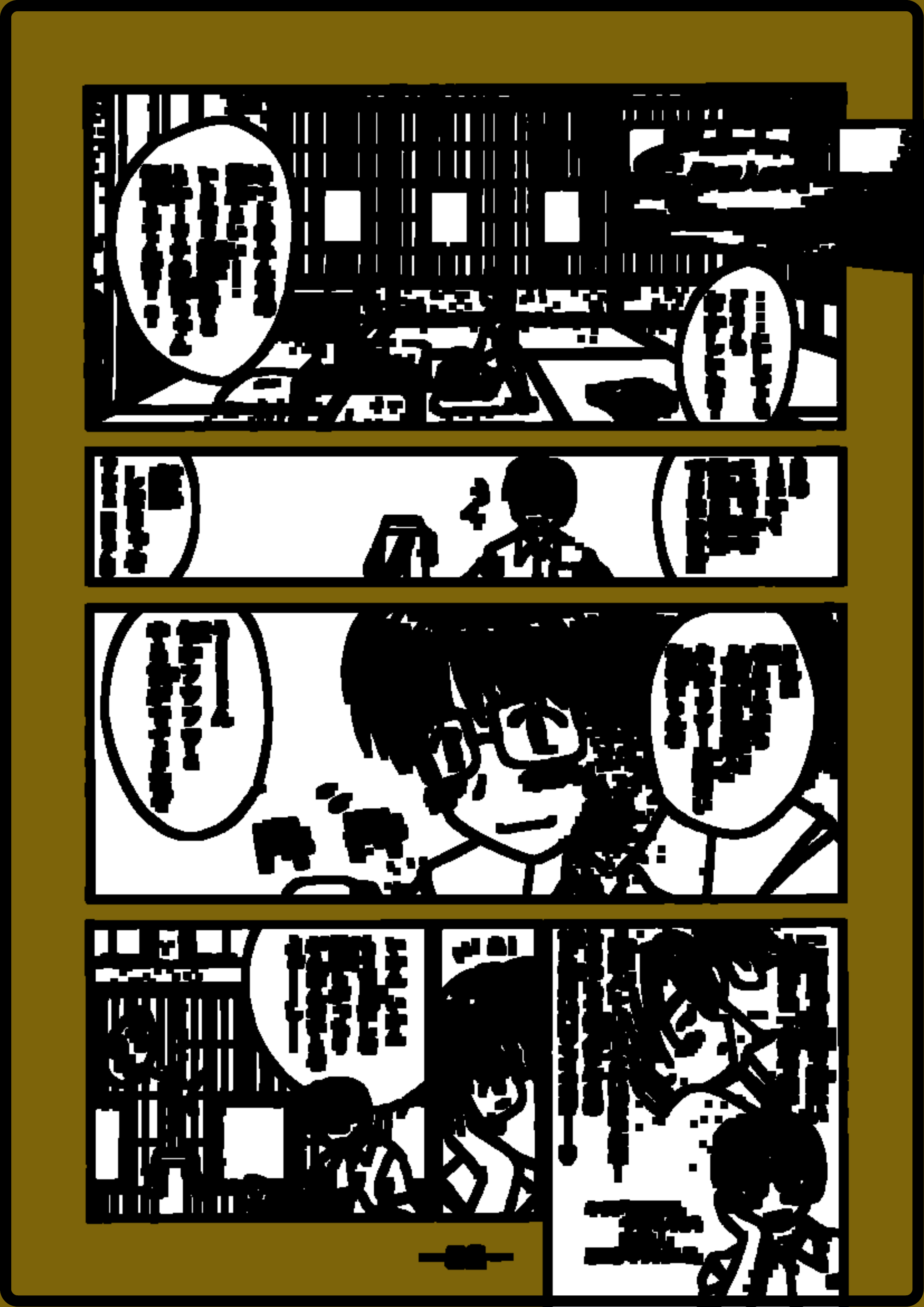}}
	\caption{(a) An input page. (b) Erosion is applied to white regions to thicken the lines. (c) White-connected areas are labeled with the same value. (d) The margin areas are selected.}
	\label{fig:framesep}
\end{figure}

\begin{figure}
	\centering
	\includegraphics[width=0.9\linewidth]{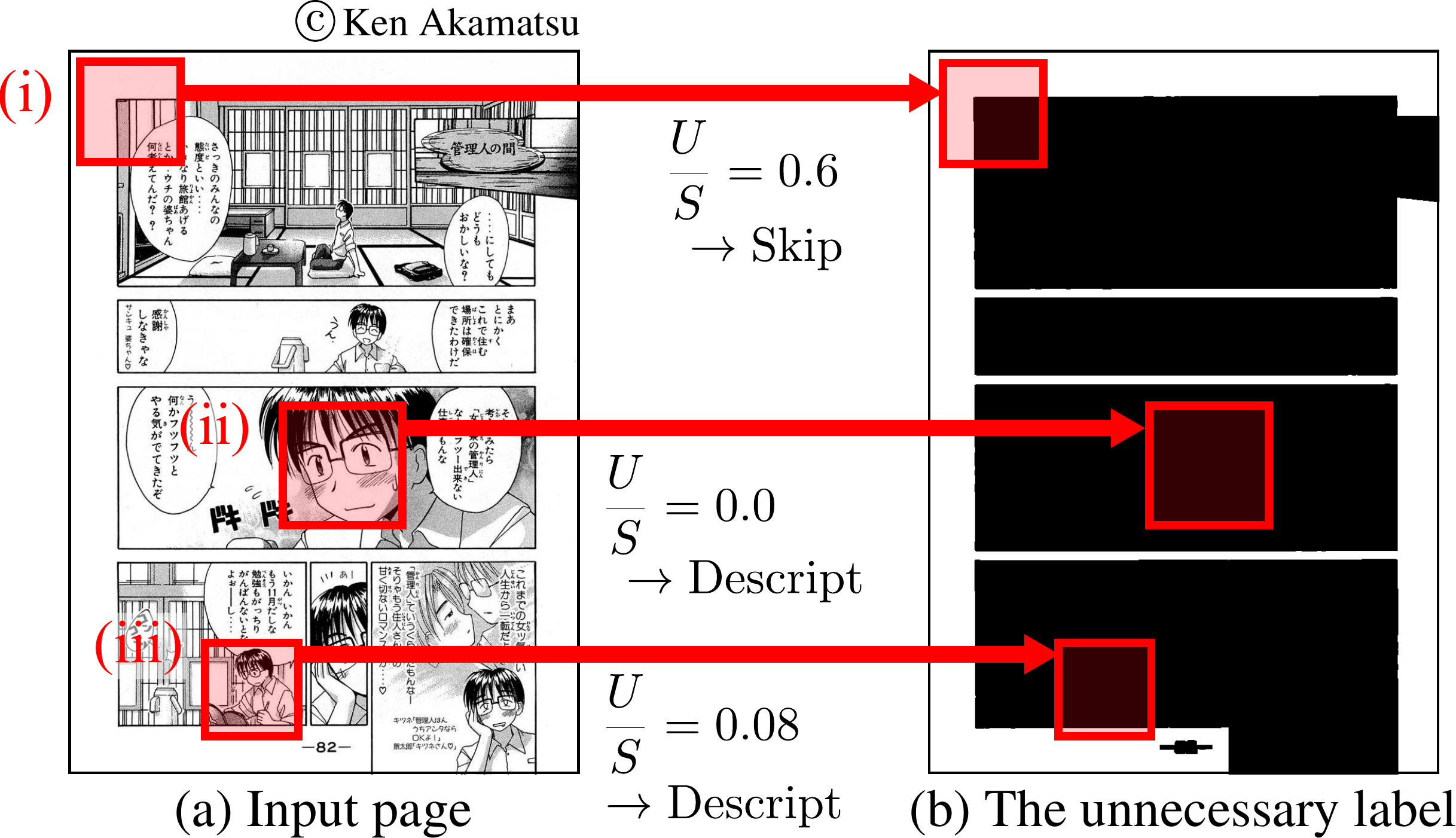} 
	\caption{(a) Input image. (b) Its margin labels. In case (i), the red area (i) in (a) is skipped because $U/S=0.6>0.1$. In case (ii), in contrast, the corresponding area is all black, and the feature is therefore extracted. In case (iii), $U/S=0.08<0.1$, and an EOH feature is extracted.}
	\label{fig:skipblank}
\end{figure}

Although PQ computing is fast, effective preprocessing of the manga pages of the dataset is essential (\Fref{fig:framework}(d)). A manga page comprises a series of frames and the interframe spaces (margins) are not important. For efficient searching, margin exclusion from retrieval candidates should be performed before extracting features. We label the margins as shown in \Fref{fig:framesep}. First, the lines of the manga page image (\Fref{fig:framesep_original}) are thickened by applying an erosion \cite{serra1983} to the white areas (\Fref{fig:framesep}(b)), thereby filling small gaps between black areas. Next, the white-connected areas are labeled by connected-component labeling \cite{marchand1999} as shown in \Fref{fig:framesep}(c), where different colors for the labels have been used for visualization. Finally, areas are selected as margins by finding the most frequent label appearing in the outermost peripheral regions (\Fref{fig:framesep}(d)). Because interframe spaces tend to connect to the outer areas, the method succeeds in most cases.

Let us define the patch area as $S(\bvec{x})$, and the margins as $U(\bvec{x})$ (e.g., the colored areas shown in \Fref{fig:framesep}(d)). We extract a feature only if its area ratio $U/S$ is less than a threshold, which we set to $0.1$. \Eref{eq:p_pq} is therefore rewritten as
\begin{equation}
\mathcal{P}_{PQ}=\left \{ \bvec{q}(\bvec{x}_i) \middle | \frac{U(\bvec{x}_i)}{S(\bvec{x}_i)} < 0.1,~~i \in \{1, \dots, N \} \right \}.
\end{equation}
Intuitively, this means that if an area belongs to the margin, it is skipped in the feature extraction step. An example of the process is shown in \Fref{fig:skipblank}.

\section{Query interaction}
\label{sec:query_interaction}

\begin{figure}
	\centering
	\subfigure[]{\label{fig:relevance_feedback_explain1} \includegraphics[width=1.0\linewidth]{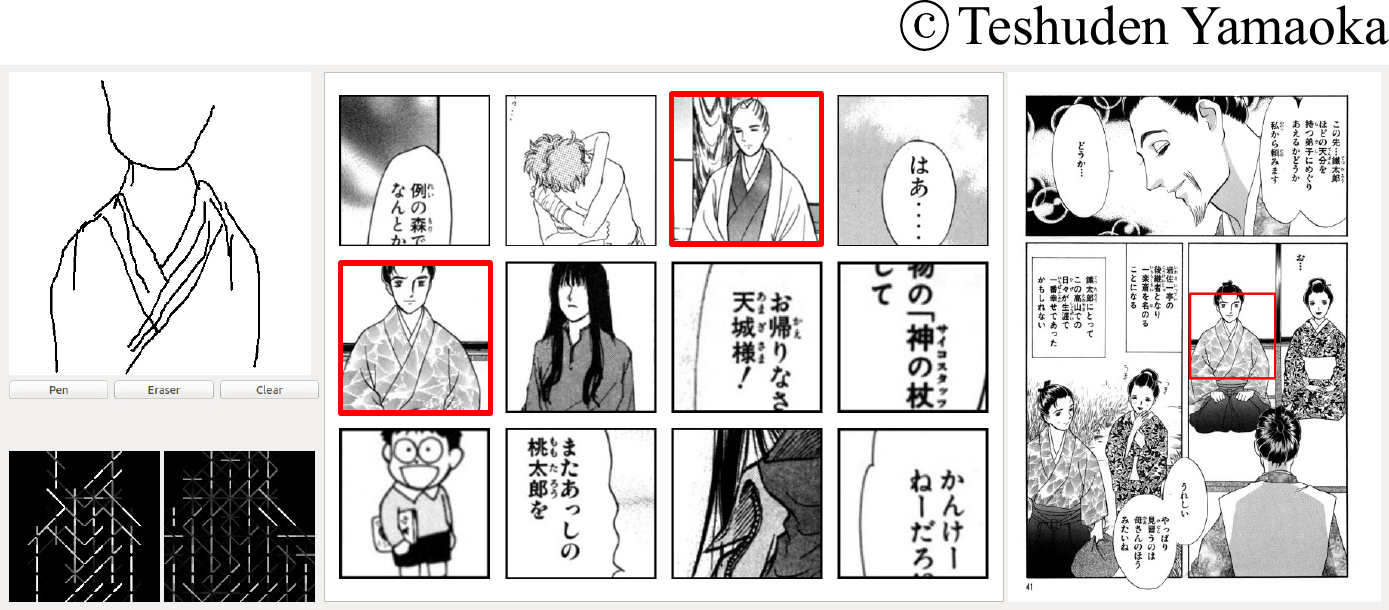}}
	\subfigure[]{\label{fig:relevance_feedback_explain2} \includegraphics[width=1.0\linewidth]{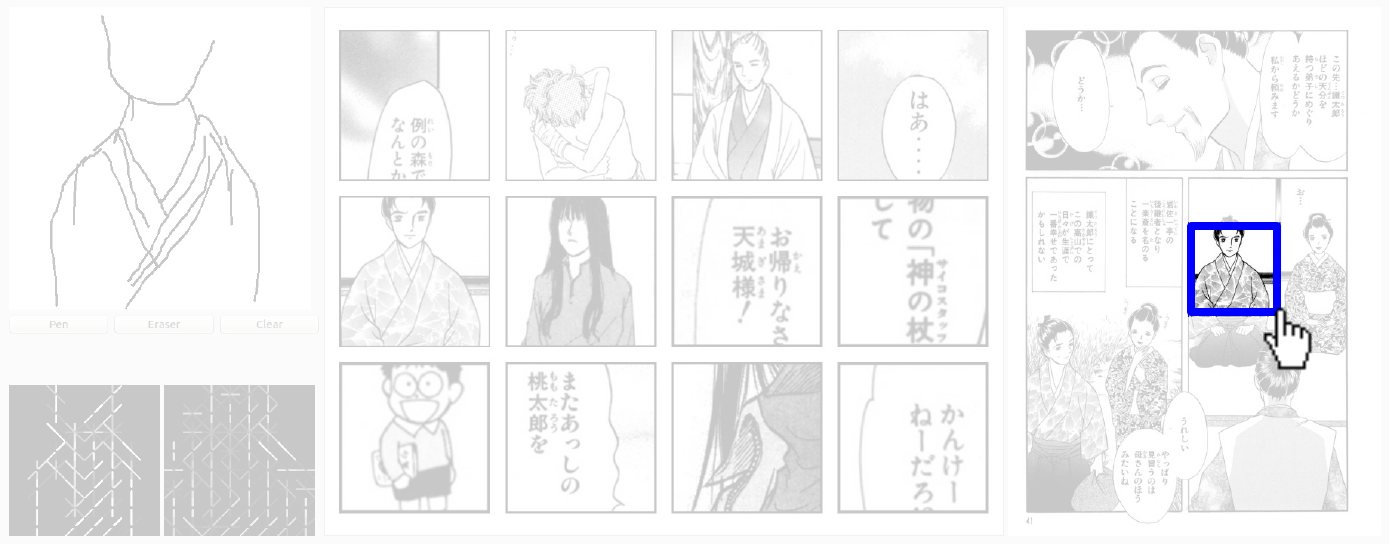}}
	\subfigure[]{\label{fig:relevance_feedback_explain3} \includegraphics[width=1.0\linewidth]{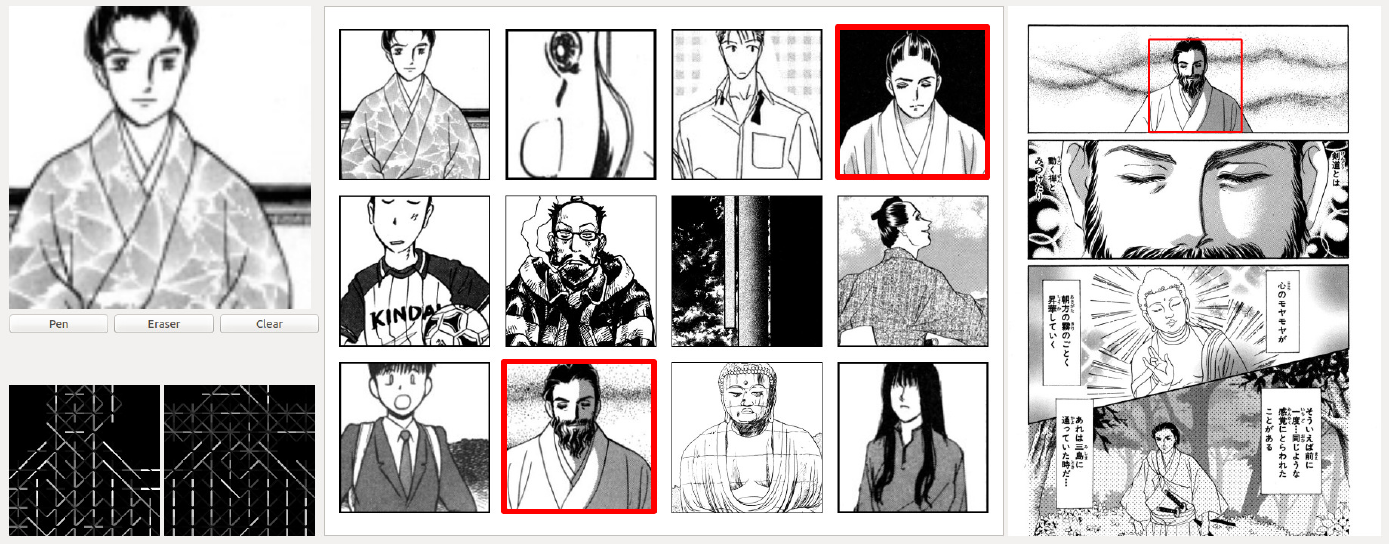}}
	\caption{Relevance feedback. (a) A user draws strokes and finds a target (Japanese-style clothes) in the third and fifth results (red borders). (b) By selecting the region in the retrieved page, another retrieval is performed automatically with this as the query. (c) A wide range of images of Japanese-style clothes is obtained.}
	\label{fig:relevance_feedback_explain}
\end{figure}

\begin{figure}
	\centering
	\subfigure[]{\label{fig:query_retouch_explain1} \includegraphics[width=1.0\linewidth]{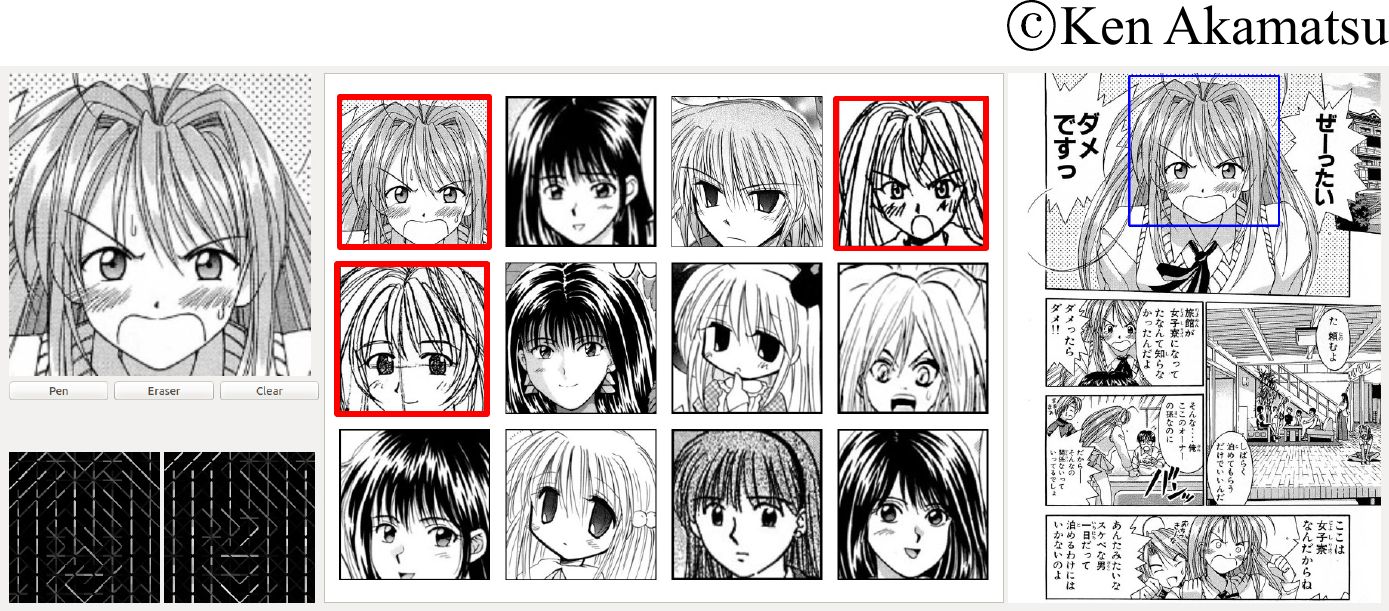}}
	\subfigure[]{\label{fig:query_retouch_explain2} \includegraphics[width=1.0\linewidth]{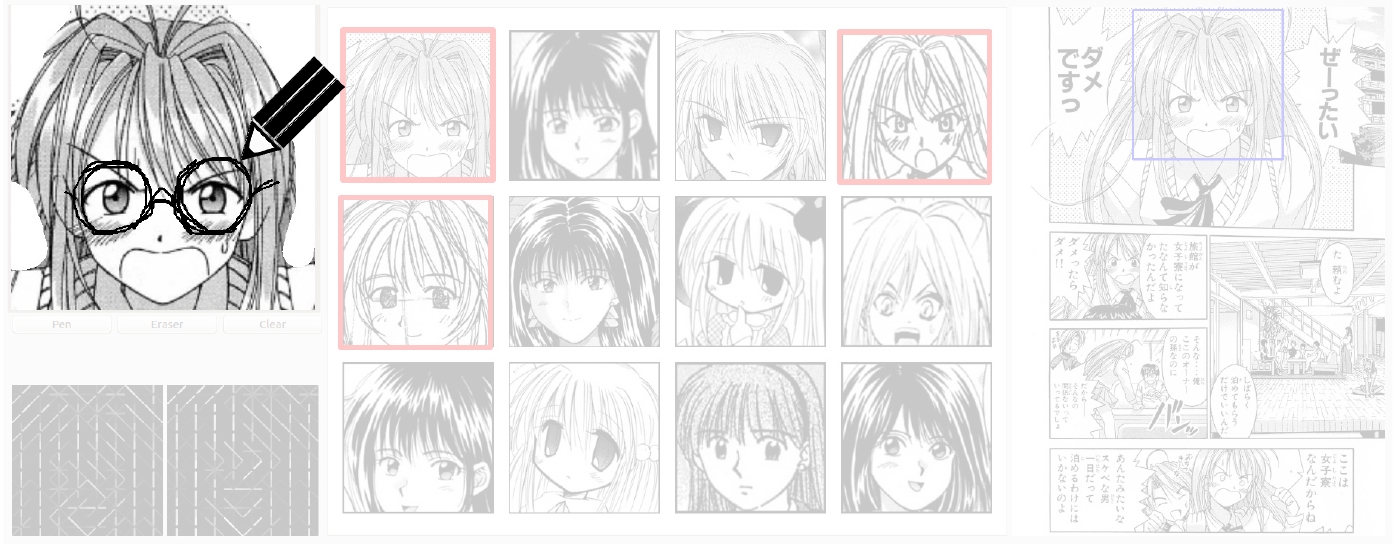}}
	\subfigure[]{\label{fig:query_retouch_explain3} \includegraphics[width=1.0\linewidth]{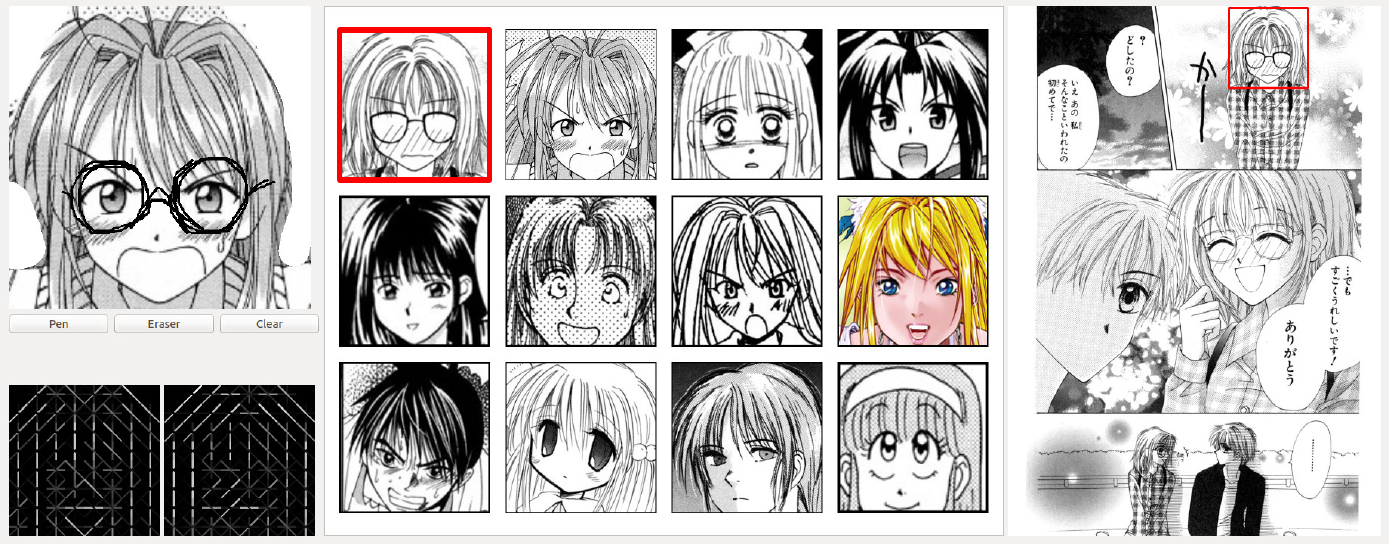}}
	\caption{Query retouch. (a) Results of relevance feedback. Target characters have red borders. (b) The user adds strokes and draws glasses on the target character. Other target characters with glasses are then successfully retrieved (red borders). Note that users can erase lines using an eraser tool. (c) Both relevance feedback and query retouch can be conducted multiple times.}
	\label{fig:query_retouch_explain}
\end{figure}

Querying is a difficult issue for manga retrieval. We call this the query modality problem of manga retrieval. We cannot employ textual or tag data because EOH features do not involve such information. For a natural and intuitive interface, we prefer sketch-based queries. Because manga itself comprise sketches drawn by authors, sketching is compatible with manga. 

In addition, we can make use of sketching not only for the initial query, but also for additional interaction with the retrieved results. The queries that can be performed using our framework are summarized as follows: (1) \textbf{sketch querying}, the proposed sketch-based retrieval described above; (2) \textbf{relevance feedback}, which reuses the retrieved results; and (3) \textbf{query retouch}, in which the results of relevance feedback are modified and reused.

\textbf{Relevance feedback}: We propose a simple interaction to reuse retrieved results, which is inspired by the relevance feedback techniques proposed in the information retrieval literature \cite{mei2014}. Users can reuse a retrieved result simply by selecting a region in the retrieved manga page, as shown in \Fref{fig:relevance_feedback_explain}. With relevance feedback, even novice users can use professional manga images as queries. Note that a user can use any region in the page as a query. 

\textbf{Query retouch}: We propose a new methodology for modifying and reusing the results. In query retouch, we can modify either the initial sketch or a query taken from the results of a retrieval (\Fref{fig:query_retouch_explain}). As the query is changed by adding lines or partial erasure, the results will change immediately. As a result, we can intuitively change the retrieval results in the direction that we want.

According to Mei and colleagues~\cite{mei2014}, these interactions are classified as ``interactive reranking,'' and are especially useful for the visual search problem because
``\textit{visual data is particularly suited for interaction, as a user can quickly grasp the vivid visual information and thus judge
the relevance at a quick glance.}''~\cite{mei2014}.

\section{Manga dataset}
\label{sec:db}
\begin{figure*}
	\centering
	\subfigure[Gakuen Noise, vol.1, p.117. \copyright Daisuke Inohara]{\label{fig:manga109_example_1} \fbox{\includegraphics[width=0.17\linewidth]{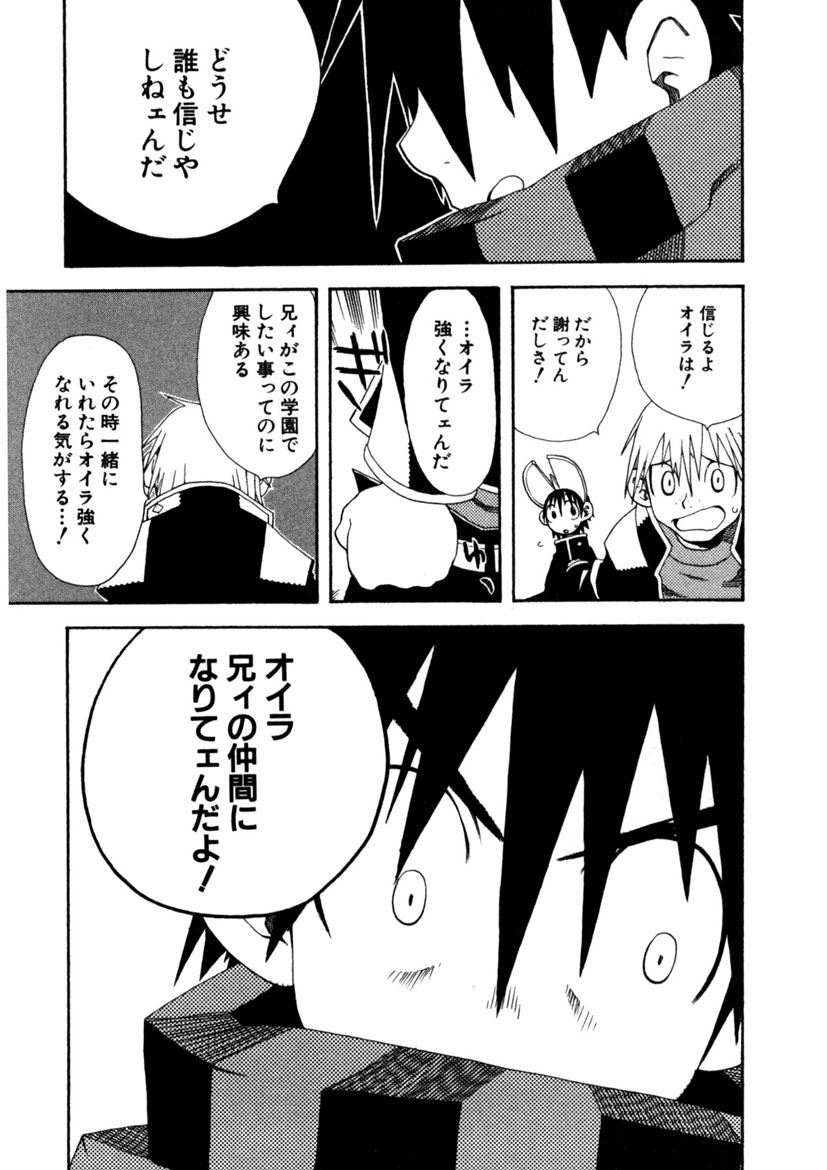}}}
	\subfigure[PLANET7, vol.2, p.36. \copyright Shuji Takeya]{\label{fig:manga109_example_2} \fbox{\includegraphics[width=0.17\linewidth]{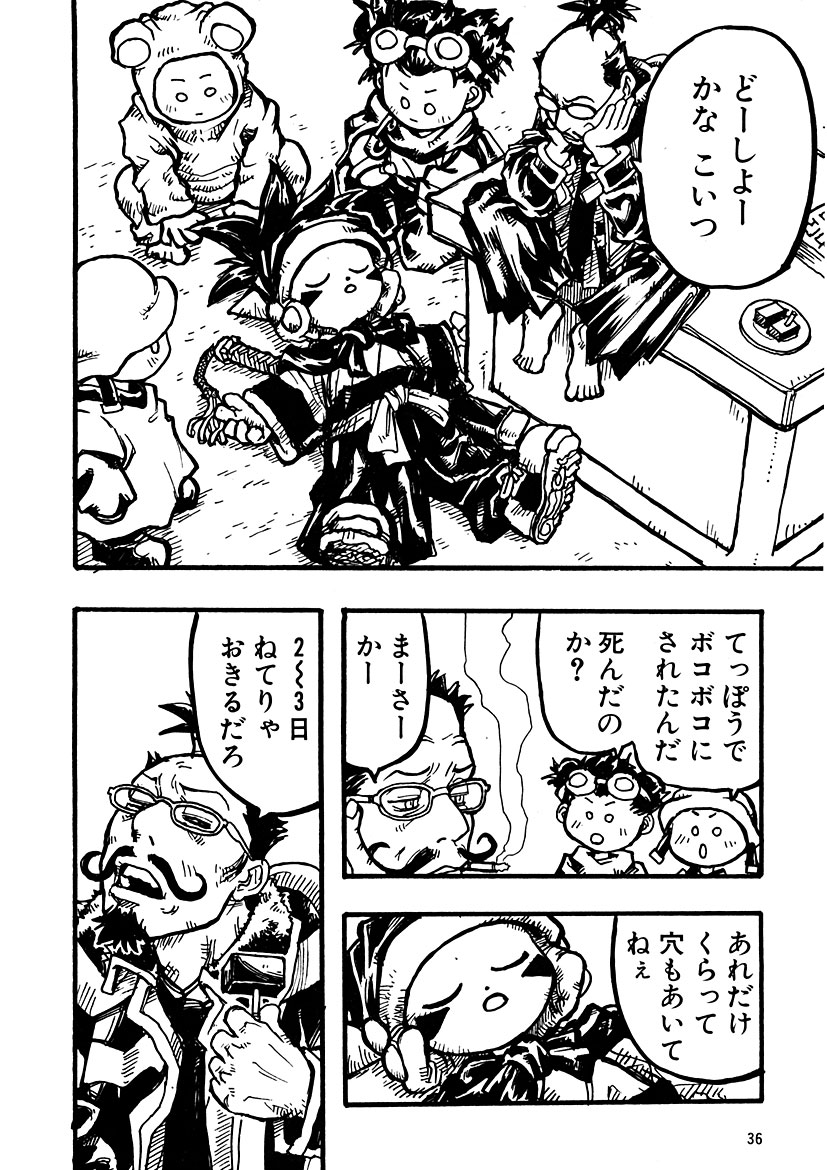}}}
	\subfigure[Aosugiru Haru. p.26. \copyright Momoko Okuda]{\label{fig:manga109_example_3} \fbox{\includegraphics[width=0.17\linewidth]{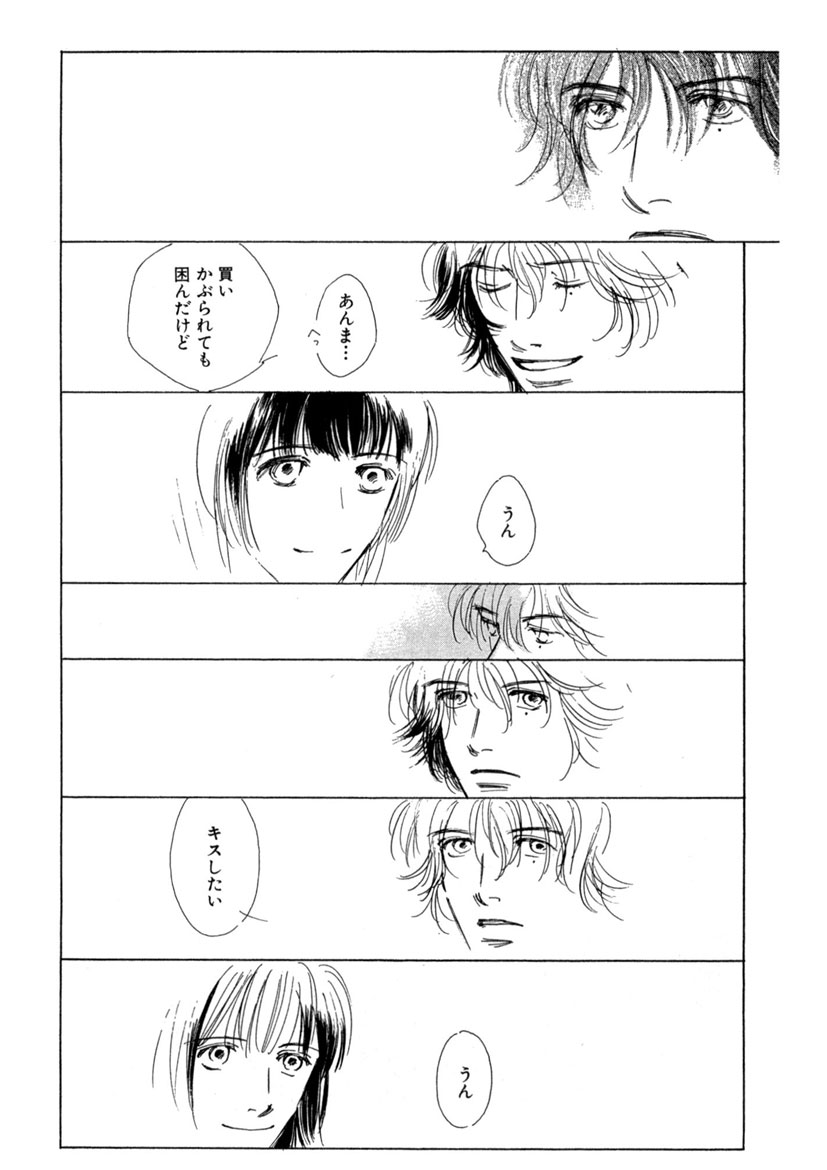}}}
	\subfigure[\scriptsize{Hanzai Kousyounin Minegishi Eitarou, vol.1, p.17. \copyright Takashi Ki}]{\label{fig:manga109_example_4} \fbox{\includegraphics[width=0.17\linewidth]{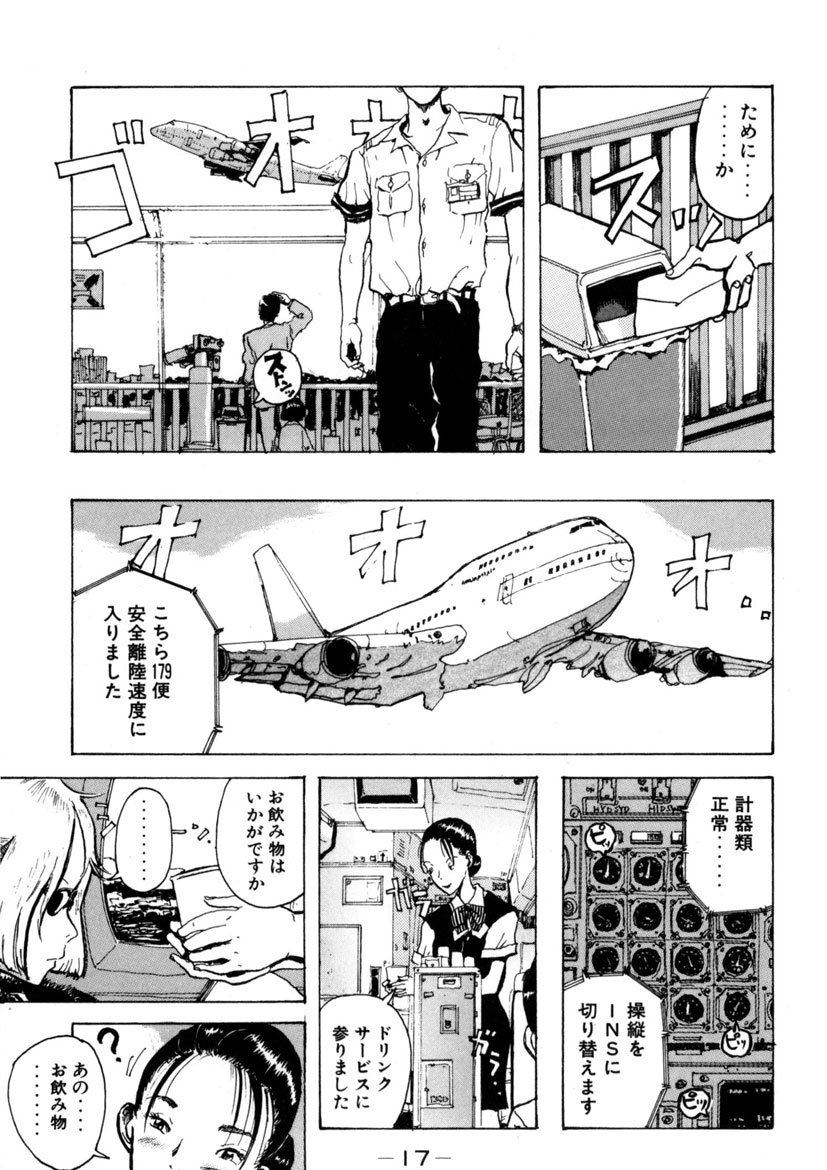}}}
	\subfigure[OL Lunch, p.18. \copyright Youko Sanri]{\label{fig:manga109_example_5} \fbox{\includegraphics[width=0.17\linewidth]{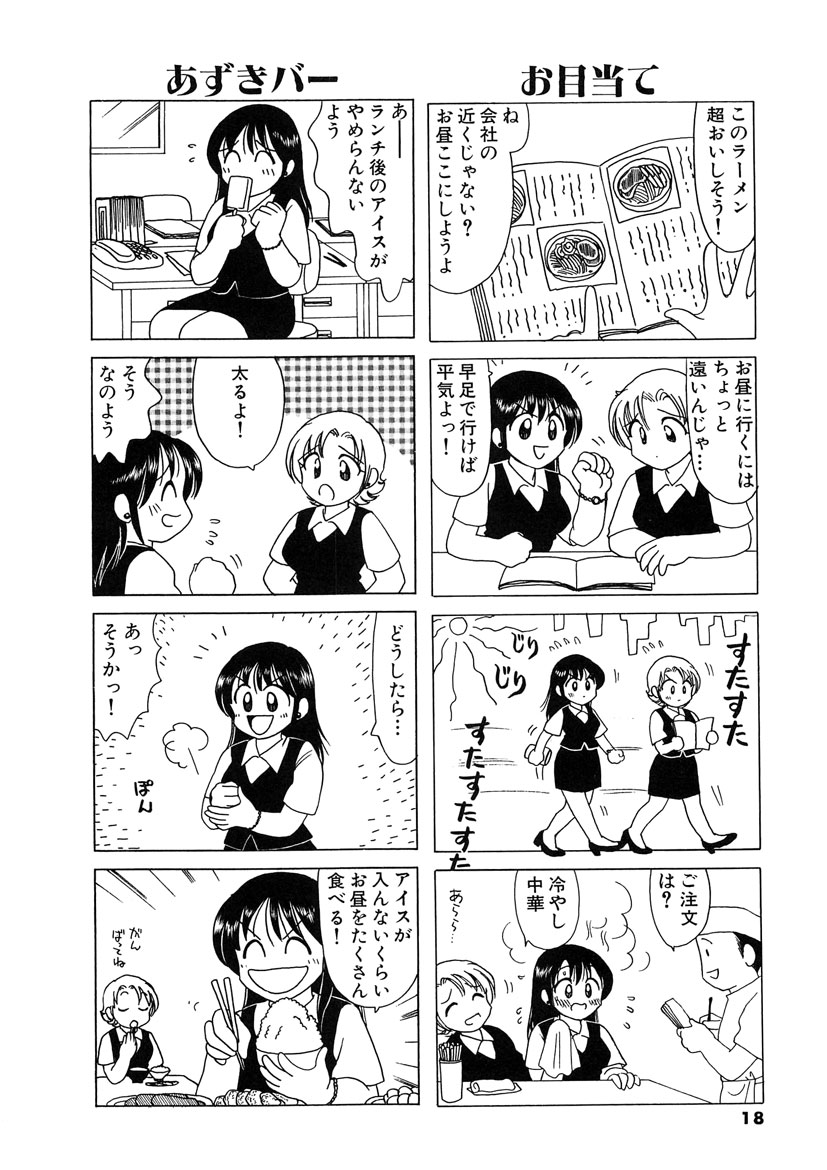}}}
	\caption{Example images from the Manga109 dataset. Each caption shows the bibliographic information of an image (title, volume, and page).}
	\label{fig:manga109_example}
\end{figure*}

In this section, we introduce a new dataset of manga images, namely the \textbf{Manga109} dataset for evaluation. The Manga109 dataset consists of 109 manga titles, and is made publicly available for academic research purposes with proper copyright notation. \Fref{fig:manga109_example} shows example pages from the Manga109 dataset. 

A large-scale dataset of manga images is important and useful for manga research. Although several papers related to manga image processing have been published, fair comparisons among the proposed methods have not been conducted because of the lack of a dataset with which the methods can be evaluated. In preparing a dataset, the most serious and intractable problem is copyright. Because manga is artwork and protected by copyright, it is hard to construct a publicly available dataset. Therefore, researchers have collected their own manga images or used manga images drawn by amateurs. These are not appropriate for manga research because: (1) the number of manga images is small; (2) the artistic quality of the manga images is not always high; and (3) they are not publicly available. The Manga109 dataset, on the other hand, clears these three conditions. It contains 21,142 pages in total, and we can say that it is currently the largest manga image dataset. The quality of the images is high because all titles were drawn by professional manga authors. 

As is widely known, well-crafted image datasets have played critical roles in evolving image processing technologies, e.g., the PASCAL VOC datasets~\cite{everingham2010} for image recognition in the 00s, and ImageNet~\cite{russakovsky2014} for recent rapid progress in deep architecture. We hope the Manga109 dataset will contribute to the further development of the manga research domain.

\textbf{Image collection:}
All manga titles in the Manga109 dataset have been previously published in manga magazines\footnote{Japanese manga is usually published in a manga magazine first, which contains many manga titles. Then it comes out in independent book form for each manga title.}, i.e., they were written by professional manga creators.
The manga titles are in an archive ``Manga Library Z'' run by J-comi. It has more than 4000 titles, most of which are currently out of print.
With the help of J-comi, we chose 109 titles from the archive that cover a wide range of genres and publication years.
We obtained permission from the creators to use them for academic research purposes.
Thus researchers can use them freely with appropriate citation for research challenges, including not only for retrieval and localization, but also for colorization, data mining from manga, and so on. The manga titles were originally published from the 1970s to the 2010s. The Manga109 dataset covers various kinds of categories, including humor, battle, romantic comedy, animal, science fiction, sports, historical drama, fantasy, love romance, suspense, horror, and four-frame cartoons. Please see the details (titles, author names, and so on) in our project page~\cite{manga109}.

\textbf{Dataset statistics:}
The dataset consists of 109 manga titles. Each title includes 194 pages on average, with a total of 21,142 pages. The average size of images is $833\times1179$, which is bigger than the image sizes usually used for object recognition and retrieval tasks, e.g., $482\times415$ pixels on average for ILSVRC 2013 \footnote{\url{http://www.image-net.org/challenges/LSVRC/2014/index#data}}. This is because manga pages contain
multiple frames and thus require higher resolutions.

\textbf{Relation to other datasets:}
\textit{eBDtheque}~\cite{guerin2013} is a publicly available comic dataset.
It consists of 100 comic pages with detailed meta-information such as text annotations. Compared with eBDtheque, our Manga109 dataset contains a much larger number of pages.
Manga109 is useful for large-scale experiments, whereas eBDtheque is beneficial for small but detailed evaluations such as object boundary detection.

\section{Experimental Results}
\label{sec:exp}
In this section, we present three evaluations: (1) a comparative study with previous methods for manga description (\Sref{sec:exp_comp}); (2) a localization evaluation (\Sref{sec:exp_local}); and (3) a large-scale qualitative study (\Sref{sec:exp_large}). These three experimental results verify that our system presents solutions to the corresponding manga image description problem, the retrieval--localization problem, and the query modality problem. In the comparative study and the localization evaluation, we used a single-thread implementation for a fair comparison, and employed a parallel implementation for the large-scale study.

\subsection{Comparative study}
\label{sec:exp_comp}
A comparative study was performed to evaluate how well the proposed framework can represent manga images compared with previous methods such as those introduced in \Sref{sec:related}.
We compared our proposal with a baseline (BoF with large window SIFT~\cite{schneider2014}), the state-of-the-art of BoF-based methods (FV~\cite{schneider2014}),
and the state-of-the-art of chamfer-based methods (Compact OCM~\cite{sun2013a}).
All experiments were conducted on a PC with a 2.8~GHz Intel Core i7 CPU and 32~GB RAM, using C++ implementations.

\begin{figure}
	\centering
	\subfigure[]{\label{fig:exp_sketch_lovehina_novice} \includegraphics[height=0.28\linewidth]{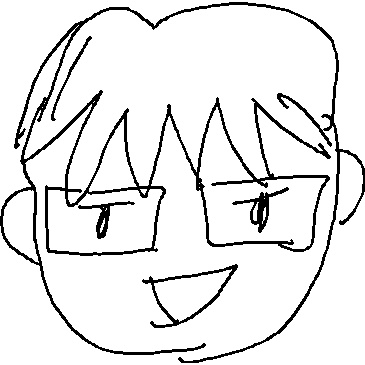}}
	\subfigure[]{\label{fig:exp_sketch_lovehina_pro} \includegraphics[height=0.28\linewidth]{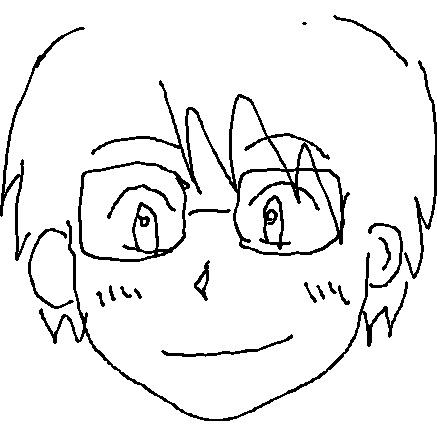}}
	\subfigure[]{\label{fig:exp_sketch_lovehina_ans} \includegraphics[height=0.28\linewidth]{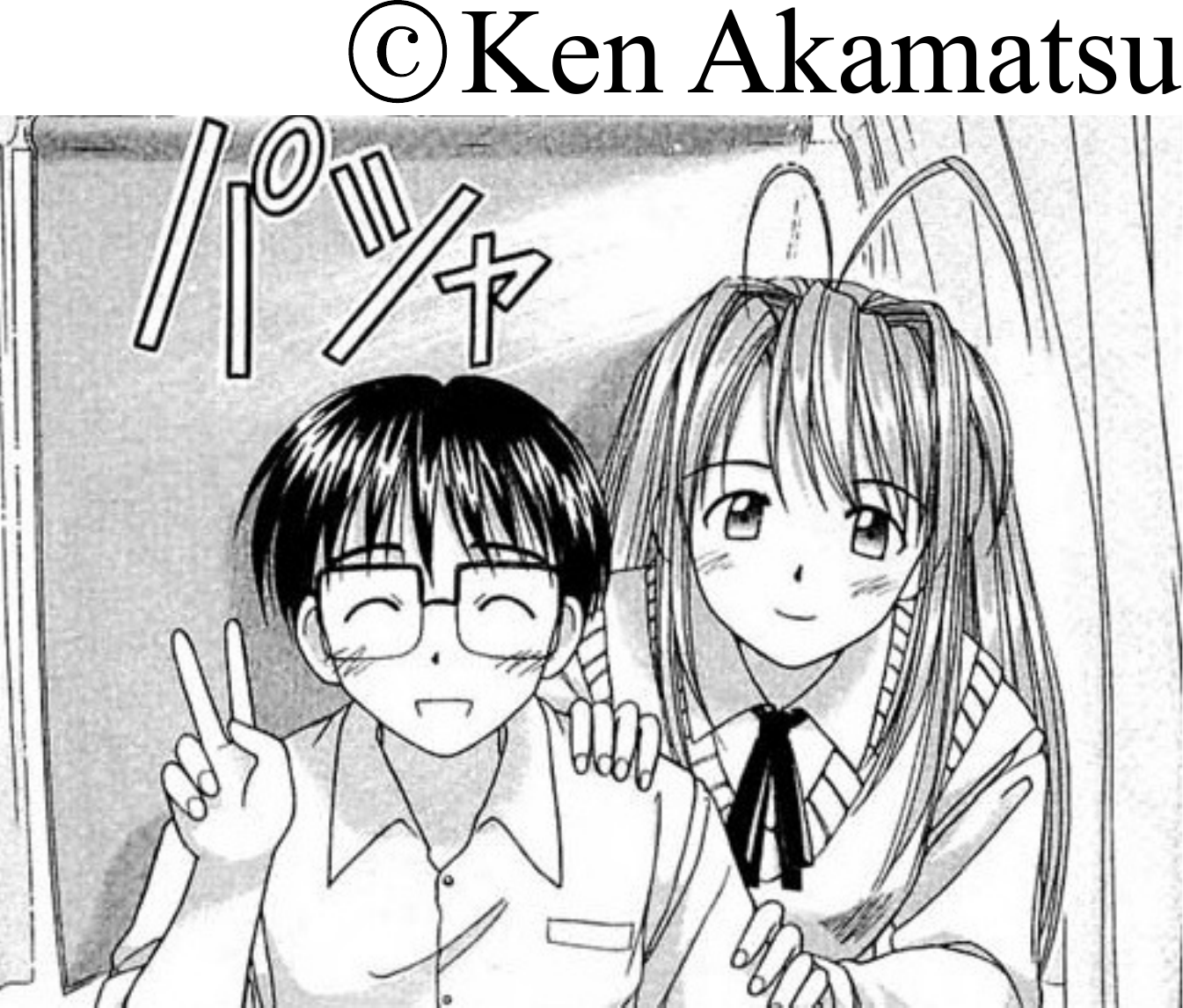}}
	\subfigure[]{\label{fig:exp_sketch_chombo_novice} \includegraphics[height=0.3\linewidth]{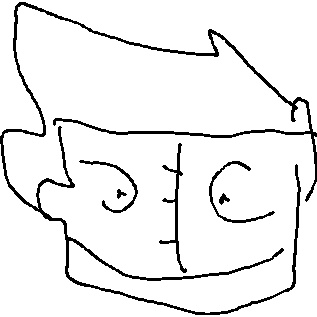}}
	\subfigure[]{\label{fig:exp_sketch_chombo_pro} \includegraphics[height=0.3\linewidth]{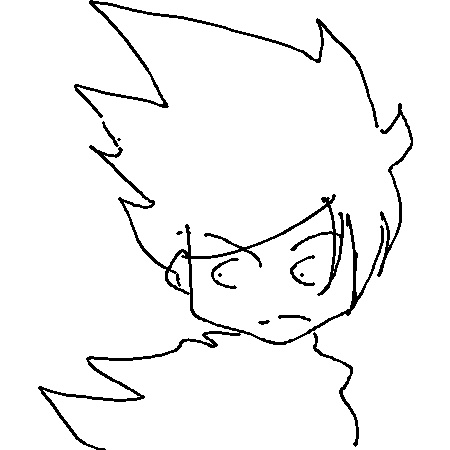}}
	\subfigure[]{\label{fig:exp_sketch_chombo_ans} \includegraphics[height=0.3\linewidth]{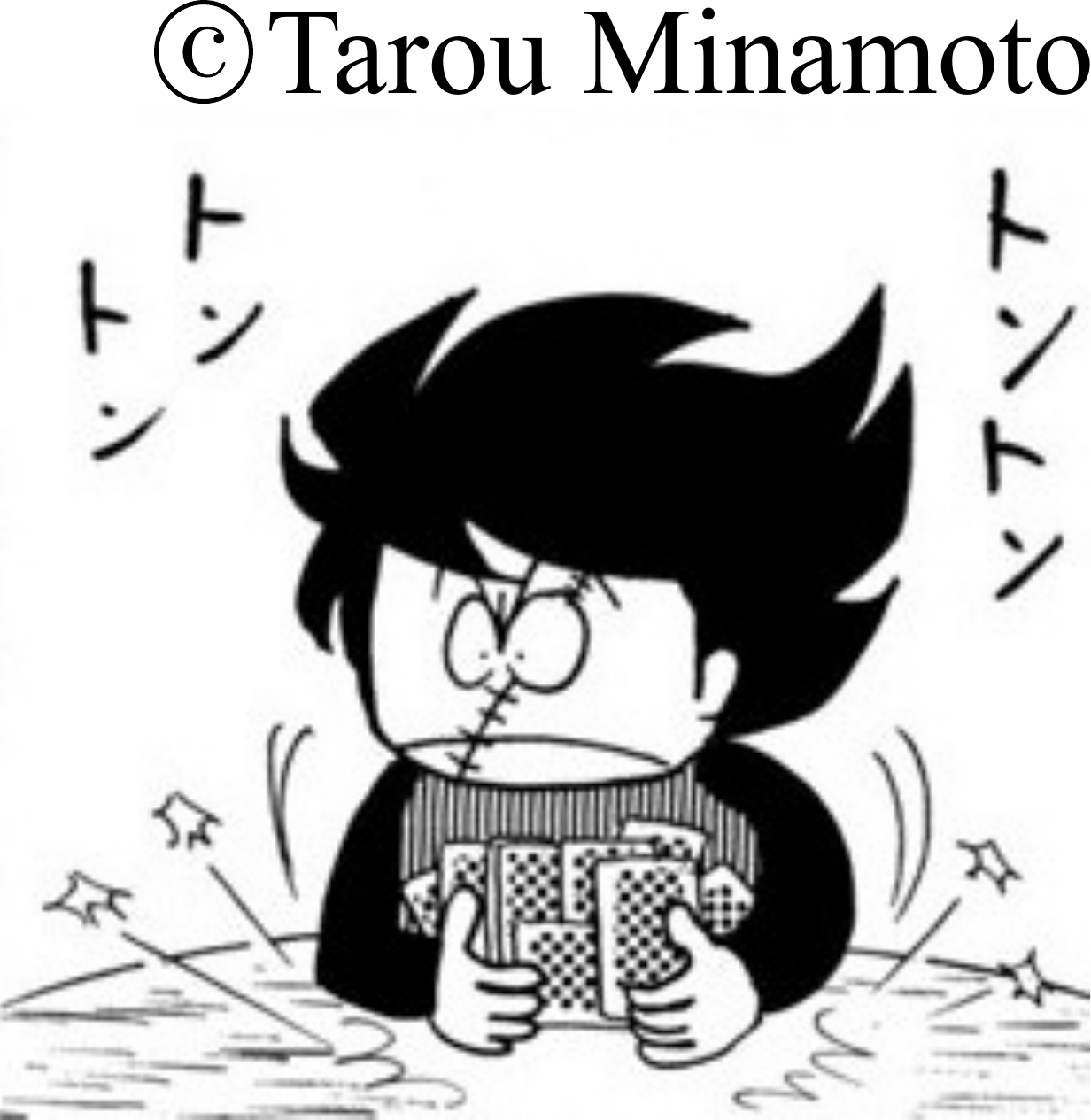}}
	\subfigure[]{\label{fig:exp_sketch_dollgun_novice} \includegraphics[height=0.3\linewidth]{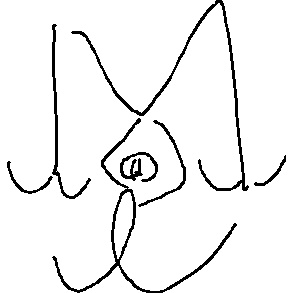}}
	\subfigure[]{\label{fig:exp_sketch_dollgun_pro} \includegraphics[height=0.3\linewidth]{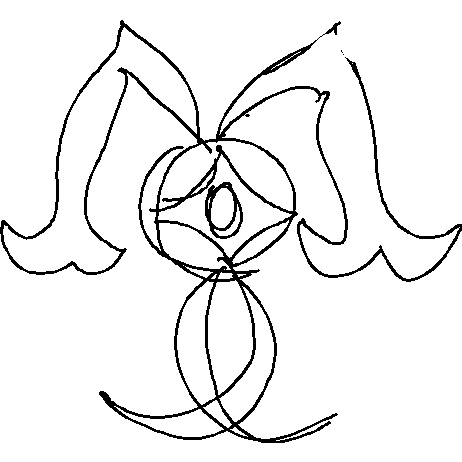}}
	\subfigure[]{\label{fig:exp_sketch_dollgun_ans} \includegraphics[height=0.3\linewidth]{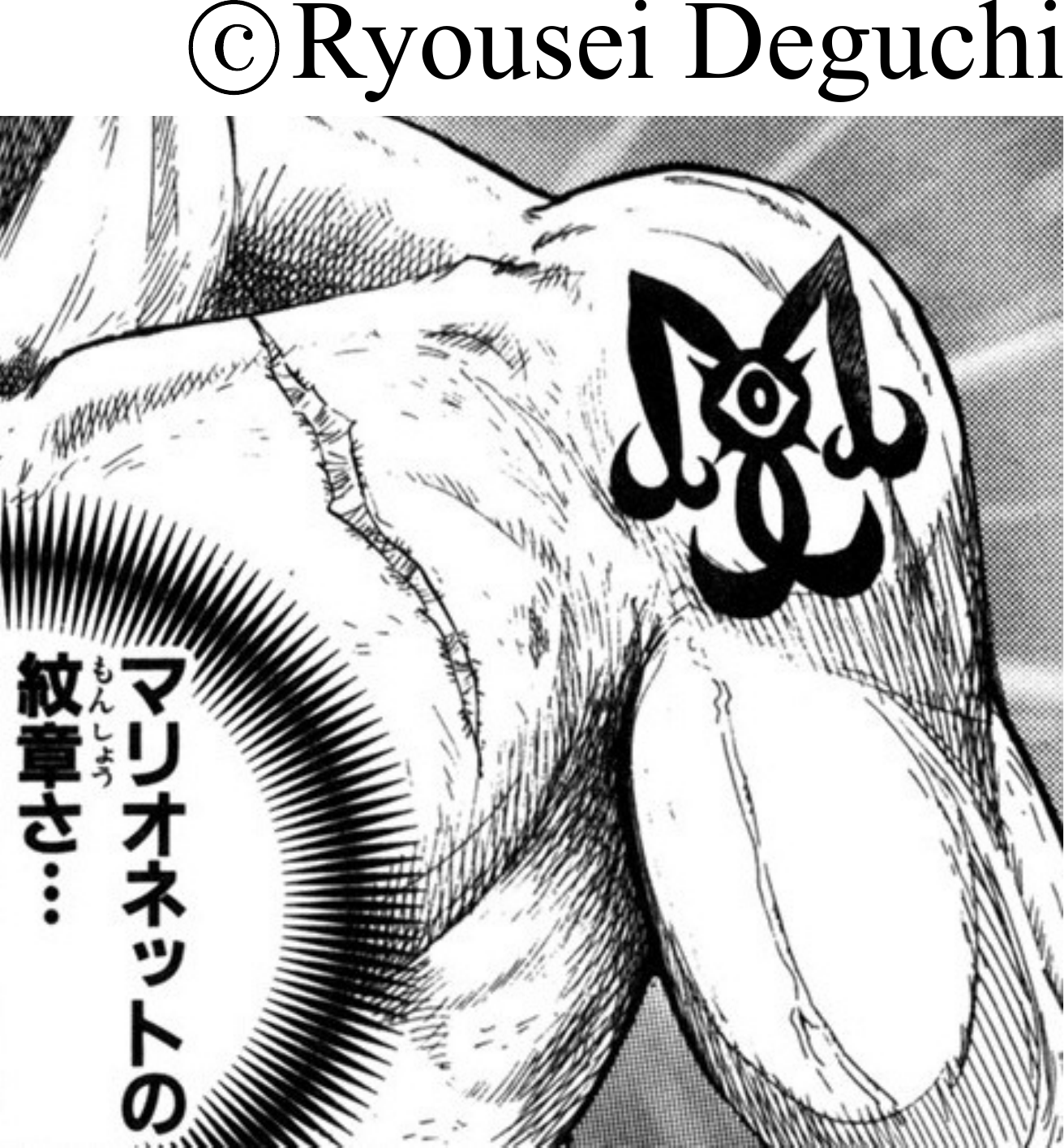}}
	\caption{Targets for the comparative study. Left to right: query sketches by novice artists, skilled artists, and ground-truth images. Top to bottom: targets, \textit{Boy-with-glasses}, \textit{Chombo}, and \textit{Tatoo}.}
	\label{fig:exp_query_answer}
\end{figure}

\textbf{Frame image dataset:}
For evaluation, we cropped frames from 10 representative manga titles from the Manga109 dataset\footnote{\textit{Lovehina vol.1}, \textit{Gakuen Noise}, \textit{DollGun}, \textit{Hanzai Kousyounin Minegishi Eitarou}, \textit{Bakuretsu KungFu Girl}, \textit{Aosugiru Haru}, \textit{OL Lunch}, \textit{Mukoukizu no Chombo}, \textit{Highschool Kimengumi vol.1}, and \textit{PLANET7}}. There were 8889 cropped frames, and the average size was $372\times341$\debug{Computed by check\_imgs\_stats.py}.
We used these frames for retrieval.
Note that we used frames instead of pages for this comparison because the features of BoF, FV, and Compact OCM are
only comparable within a frame.
Although the frame is less complex than the page, retrieval is 
not easy because the size of frames varies greatly, and the localization problem still exists as shown in \Fref{fig:exp_sketch_lovehina_ans}, where the target (a head of a boy) is small and the frame includes other objects and backgrounds.

\textbf{Target images:} To make the comparisons, we chose three kinds of targets: \textit{Boy-with-glasses} (\Fref{fig:exp_sketch_lovehina_ans}), \textit{Chombo} (\Fref{fig:exp_sketch_chombo_ans}), and \textit{Tatoo} (\Fref{fig:exp_sketch_dollgun_ans}), which are from manga titles \textit{Lovehina}, \textit{Mukoukizu no Chombo}, and \textit{DollGun}, respectively. \textit{Boy-with-glasses} is an easy example, \textit{Chombo} is much harder because the face might be either to the right or left, and \textit{Tatoo} is the most difficult because it is so small compared with the frame. These target images are treated as ground truths.

\begin{table}
	\caption{Image statistics for comparative study. All images are cropped frames from the Manga109 dataset.}
	\label{tbl:statistics_comp}
	\centering
	\begin{tabular}{cccccc} \hline
		target type & difficulty & \#ground truth & \#dataset & \#query \\ \hline
		Boy-with-glasses & easy & 67 & 8563 & 30 \\
		Chombo & normal & 178 & 8563 & 30 \\
		Tatoo & hard & 81 & 8563 & 30 \\ \hline
	\end{tabular}
\end{table}

\textbf{Query images:}
To prepare query sketches, we invited 10 participants (seven novices and three skilled artists, such as a member of an art club), and asked them to draw sketches. First, we showed them query images for 20 s. Next, they were asked to draw a sketch for the query. Each participant drew all target objects; therefore, we collected $10\times3=30$ queries. Examples of queries are shown in \Fref{fig:exp_query_answer}. We used a 21-inch pen display (WACOM DTZ-2100) for drawing.

\textbf{Results of comparative study:}
Using the queries, we evaluated each method using standard evaluation protocols in image retrieval: recall@k and mean average precision~\cite{buttcher2010}. Note that the ground-truth target images were labelled manually. All frame images were used for the evaluation, and the images of the different targets were regarded as distractors. The statistics of the frame images are shown in \Tref{tbl:statistics_comp}.

\begin{figure*}
	\centering
	\subfigure[Boy-with-glasses]{\label{fig:exp_comp_lovehina} \includegraphics[width=0.32\linewidth]{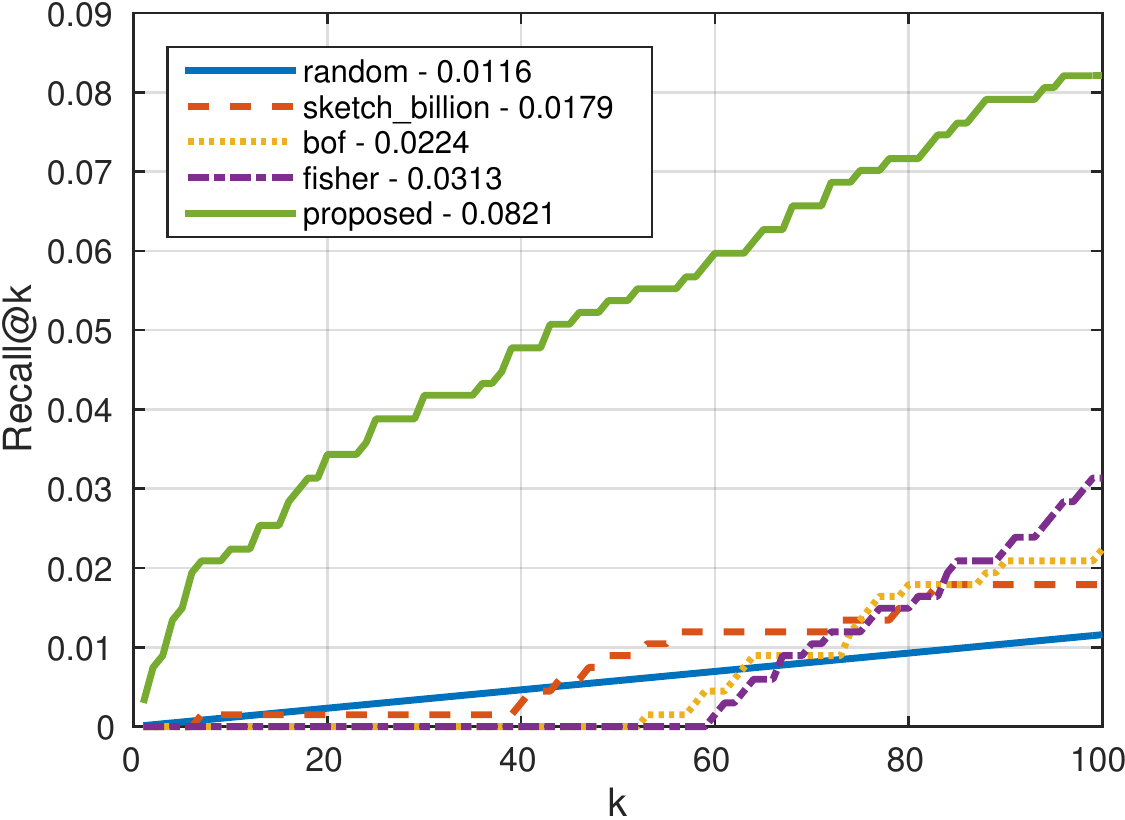}}
	\subfigure[Chombo]{\label{fig:exp_comp_chombo} \includegraphics[width=0.32\linewidth]{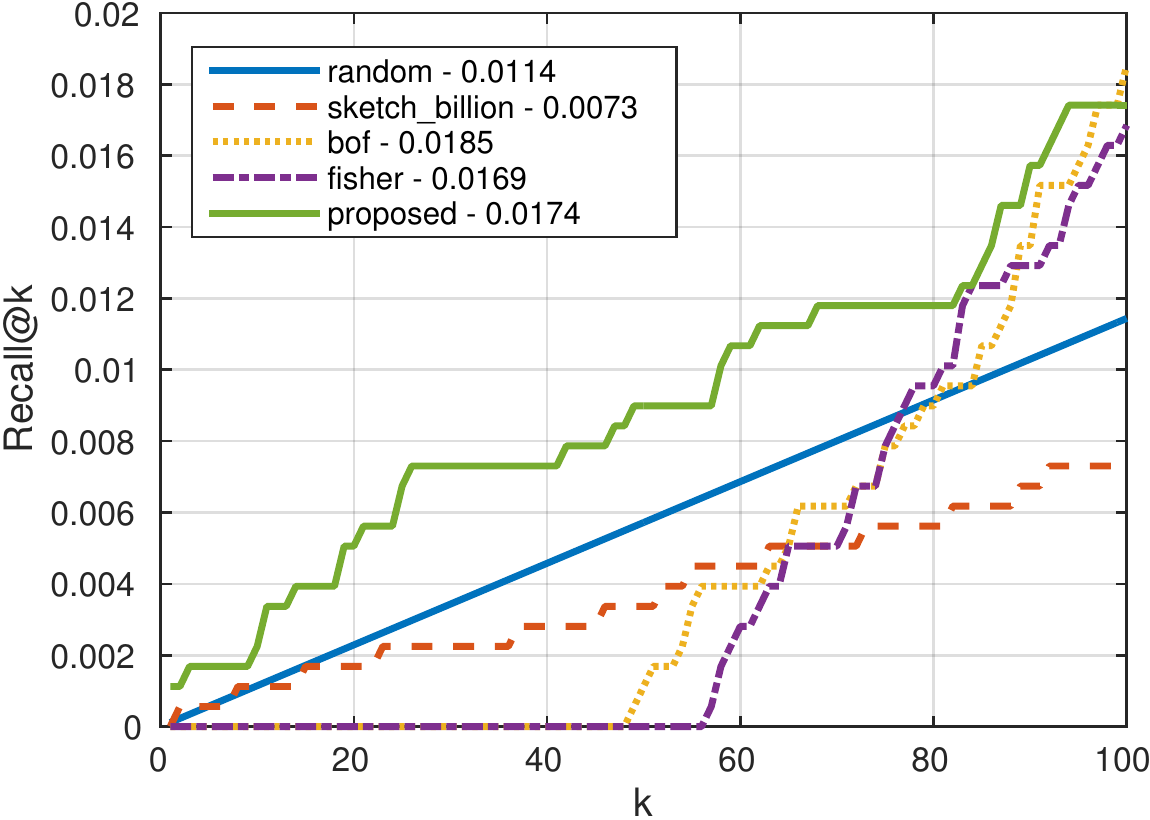}}
	\subfigure[Tatoo]{\label{fig:exp_comp_dollgun} \includegraphics[width=0.32\linewidth]{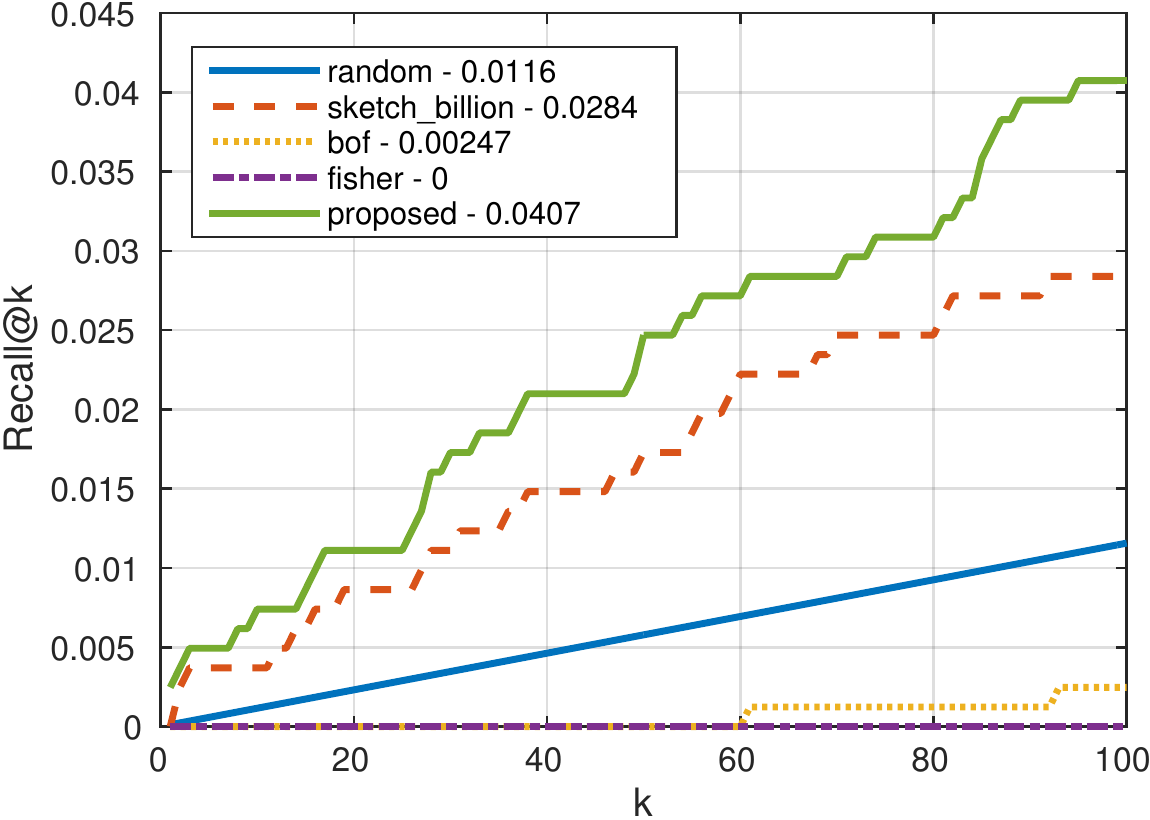}}
	\caption{Results of the comparative study. Values in the legend show Recall@100.}
	\label{fig:exp_comp}
\end{figure*}

\Fref{fig:exp_comp} shows the results for each target. Note that this task is challenging, and all scores tend to be small. The queries from users are not always similar to the target. On the contrary, some novices even drew queries that were dissimilar to the target, as shown in \Fref{fig:exp_sketch_chombo_novice} and \Fref{fig:exp_sketch_chombo_ans}. Still, in all cases, the proposed method achieved the best scores. In particular, the proposed method outperformed the other methods for \textit{Boy-with-glasses}. Note that BoF and FV received almost zero scores for the \textit{Tatoo} case because they are not good at finding a relatively small instance from an image. From these experiments, we can say that BoF-based methods do not satisfy the purpose in manga retrieval.

Note that we did not apply any approximation steps for a fair comparison. Compact OCM measures an original chamfer distance without an approximation (a sparse projection). We did not compress a feature using PQ in the proposed method.

\begin{figure}
	\centering
	\includegraphics[width=1.0\linewidth]{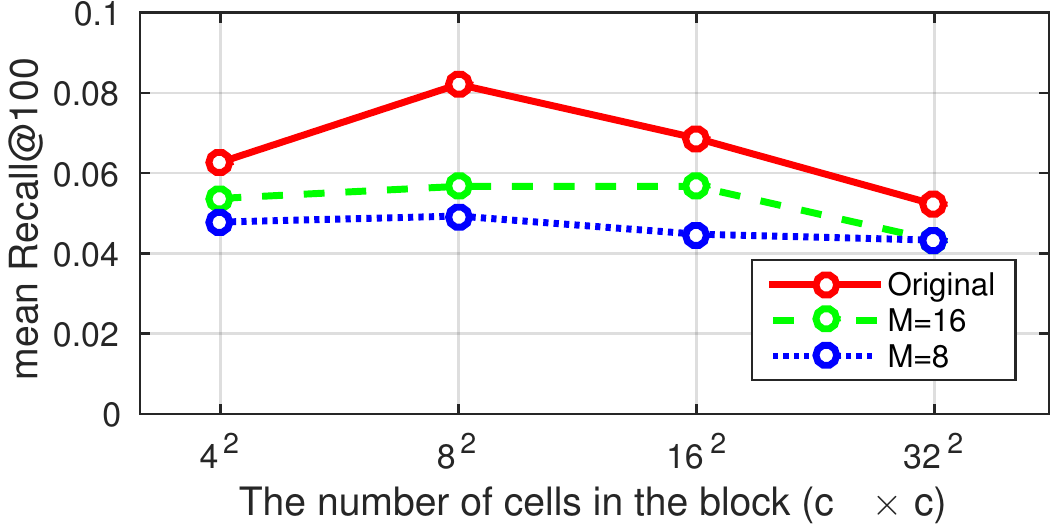}
	\caption{The effect of feature compression by PQ. The $Y$-axis represents the retrieval performance. The $X$-axis shows the number of cells. Each line corresponds to a compression level. As the features are compressed by PQ (i.e., the feature is represented by a smaller number of subvectors ($M$)), the score decreases compared with the original uncompressed feature.}
	\label{fig:exp_div}
\end{figure}

\textbf{About quantization:}
When we applied PQ to the features, the score decreased according to the quantization level, as shown in \Fref{fig:exp_div}. There is a clear trade-off between the rate of compression and accuracy. From the result, we accepted $M=16$ as a reasonable option; i.e., a feature is divided into 16 parts for PQ compression and encoded to a 16-byte code. 
Note that, interestingly, there is not a clear relation between accuracy and the number of cells ($c^2$).
The feature description becomes finer with a large number of cell divisions, but it does not always achieve higher recall.
This indicates that some level of abstraction is required for the sketch retrieval task. 
We accepted $c=8$ for all experiments, i.e., a selected area is divided into $8\times8$ areas for feature description.

\textbf{Parameter settings and implementation details:}
We show parameter settings and implementation details used for the evaluation.
For BoF, SIFT features are densely extracted where the size of a SIFT patch is $64\times 64$ pixels, with a $2\times2$ spatial pyramid, and the number of vectors in the dictionary is set to 1024. The final number of dimension is 4096. 
For FV, a Gaussian Mixture Model with 256 Gaussians was used, with a $2\times2$ spatial pyramid. The dimensions of SIFT are reduced from 128 to 80 by PCA. For BoF and FV, we leveraged the same parameter settings as in Schneider and colleagues~\cite{schneider2014}, with the \texttt{vlfeat} implementation~\cite{vedaldi2010}.
For selective search, we employed the \texttt{dlib} implementation~\cite{king2009}, with little Gaussian blurring before applying selective search. 
To train code words for BoF, FV, and PQ, we randomly selected 500 images from the dataset. Note that they were excluded and not used for testing.
To eliminate small patches, we set a minimum length of a patch as 100 pixels and discarded patches that were smaller than that.

\subsection{Localization evaluation}
\label{sec:exp_local}
Next, we evaluated how well the proposed method can localize a target object in manga pages. The setup is similar to that for image detection evaluation~\cite{everingham2010}.

\textbf{Images}:
As query sketches, we used the Boy-with-glasses sketches collected in \Sref{sec:exp_comp}. The ground-truth windows were manually annotated from Lovehina. A total of 69 windows was annotated. For evaluation, we prepared two datasets. (i) A Lovehina dataset, which is a title of \textit{Lovehina vol.1}, and consists of 192 pages, including the pages containing ground-truth windows. (ii) The Manga109 dataset, which is the sum of all manga data, consists of 109 titles with a total of 21,142 pages. Note that the Lovehina data is included in the Manga109 dataset, so (i) is a subset of (ii).

In contrast to the previous comparative study, localization is performed on a page. Therefore, this is an object localization task: given a query image, find a target instance in an image. In our case, however, we must find the target from many manga pages (21,142 pages, for Manga109).

\textbf{Evaluation criteria}
For evaluation, we employed a standard PASCAL overlap criterion~\cite{everingham2010}. Given a bounding box (retrieved result)
from the method, it is judged to be true or false by measuring
the overlap of the bounding box and ground-truth windows.
Denote the predicted bounding box as $B_p$, and the ground-truth 
bounding box as $B_{gt}$, the overlap is measured by:
\begin{equation}
\label{eq:pascal_overlap}
r = \frac{area(B_p \cap B_{gt})}{area(B_p \cup B_{gt})}.
\end{equation}
We judged the retrieved area is true if $r>0.5$.
If multiple bounding boxes are produced, at most one among them is counted
as correct.

For each query, we find the top 100 areas using the proposed retrieval method from the dataset (Lovehina or Manga109), then the retrieved
areas are judged using \Eref{eq:pascal_overlap}. Then we can compute the standard mAP@100 from the result (true/positive sequence).

\begin{figure*}
	\centering
	\includegraphics[width=1.0\linewidth]{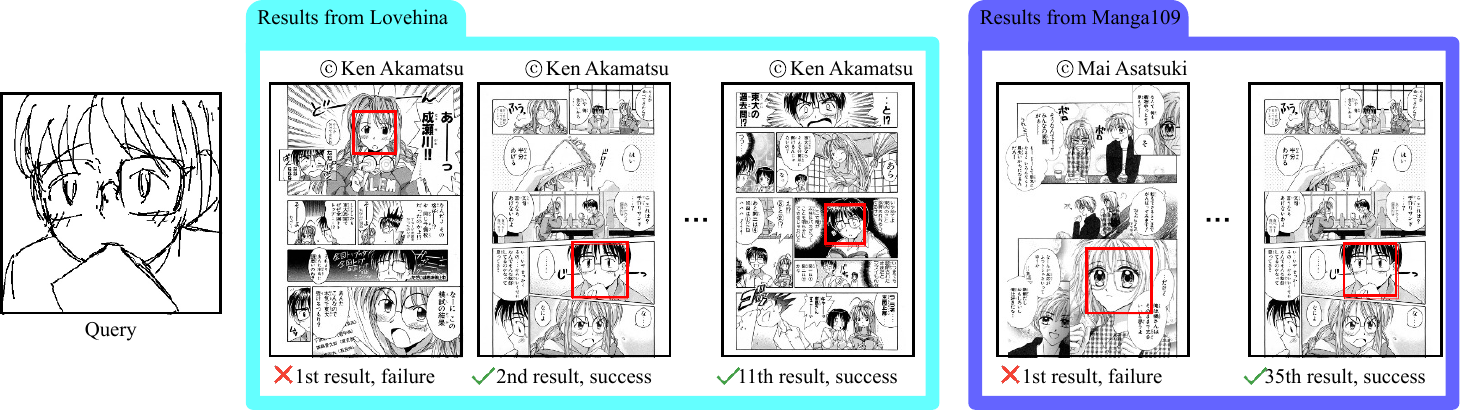}
	\caption{Examples of localization experiments for the Lovehina dataset and Manga109 dataset.}
	\label{fig:exp_localization_example}
\end{figure*}

\textbf{Result of localization evaluation}:
We show the results in \Tref{tbl:localization}.
With our single implementation, searching the Manga109 dataset (14M patches) took 331 ms. This is fast enough, and was further improved (70 ms) using a parallel implementation as discussed in \Sref{sec:exp_large}. We also show theoretical values of memory consumption for EOH features (\#path$\times 8M$). The whole Manga109 dataset consumes only 204 MB.
As can be seen from mAP, this task is difficult. There are possibly hundreds of thousands of candidate areas in the windows (138K for Lovehina, and 14M for Manga109), but only 69 ground-truth areas. Examples of retrieved results are shown in \Fref{fig:exp_localization_example}. For the Lovehina data, the first result is a failure but the second is correct. For the Manga109 dataset, the first success can be found at the 35th result. The first result shares similar characteristics to the query (wearing glasses) even though the result is incorrect.

\begin{table}
	\caption{Results for localization evaluation (single thread implementation).}
	\label{tbl:localization}
	\centering
	\begin{tabular}{cccccc} \hline
		& \#image & \#patch & mAP@100 & memory & runtime \\ \hline
		Lovehina & 192 & 138K & 1.12 $\times 10^{-2}$ & 2.21 MB & 11.6 ms \\
		Manga109 & 21,142 & 14M & 1.43 $\times 10^{-4}$ & 204 MB & 331 ms \\ \hline
	\end{tabular}
\end{table}

\subsection{Large-scale qualitative study}
\label{sec:exp_large}
In this section, we show a qualitative study of retrieval from the Manga109 dataset. The whole system was implemented using a GUI as shown in \Fref{fig:gui}.

We employed a parallel implementation using the Intel Thread Building Library, and the average computation time was 70 ms for the Manga109 dataset (21,142 images).
The parallelization was straightforward. Neighbors are computed for each manga title in parallel.
In the implementation, we selected the most similar feature from a page (not keeping all features per page), then merged the results.

\textbf{Qualitative study using a sketch dataset:}
We qualitatively evaluated the proposed method using 
a public sketch dataset as queries.
We used representative sketches~\cite{eitz2012} as queries.
The 347 sketches each had a category
name, e.g., ``panda.''

\begin{figure}
	\centering
	\subfigure[Trousers]{\label{fig:eitz_trousers} \includegraphics[width=1.0\linewidth]{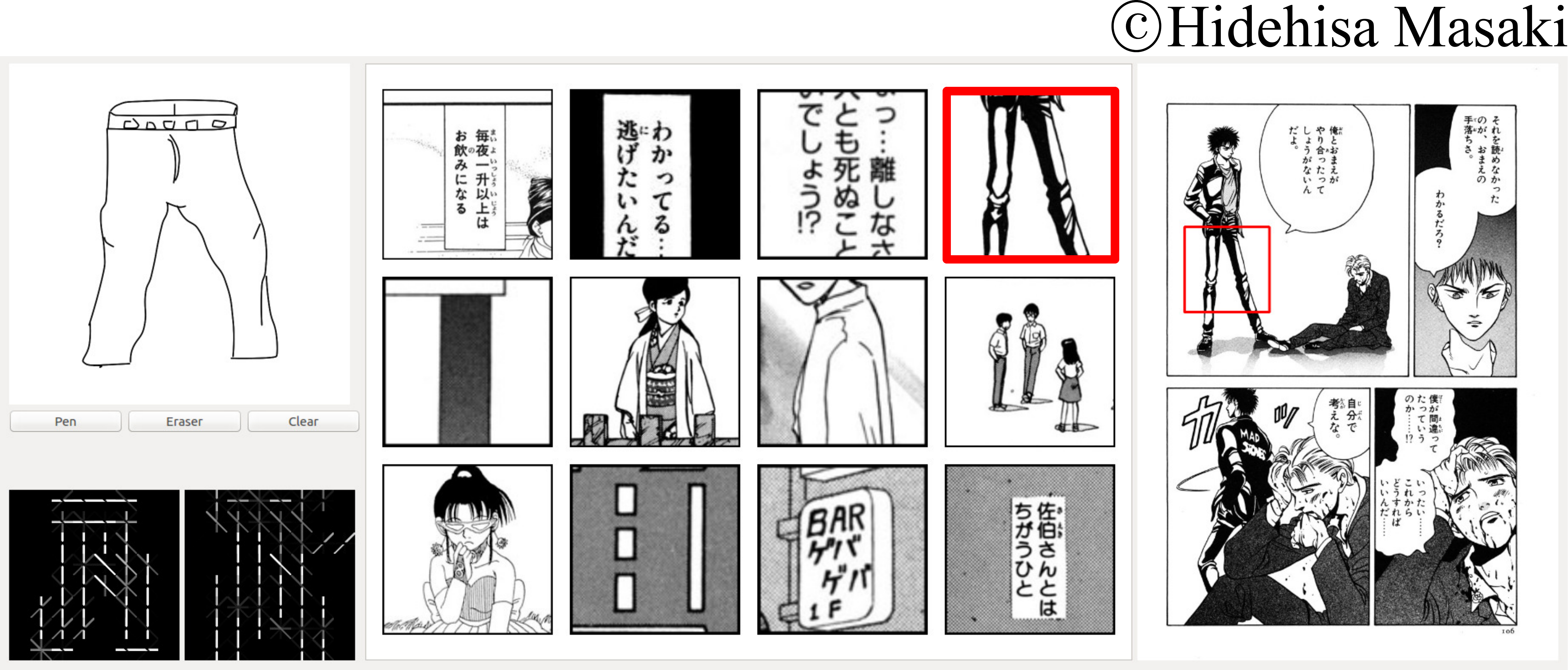}}
	\subfigure[Computer monitor]{\label{fig:eitz_computer_monitor} \includegraphics[width=1.0\linewidth]{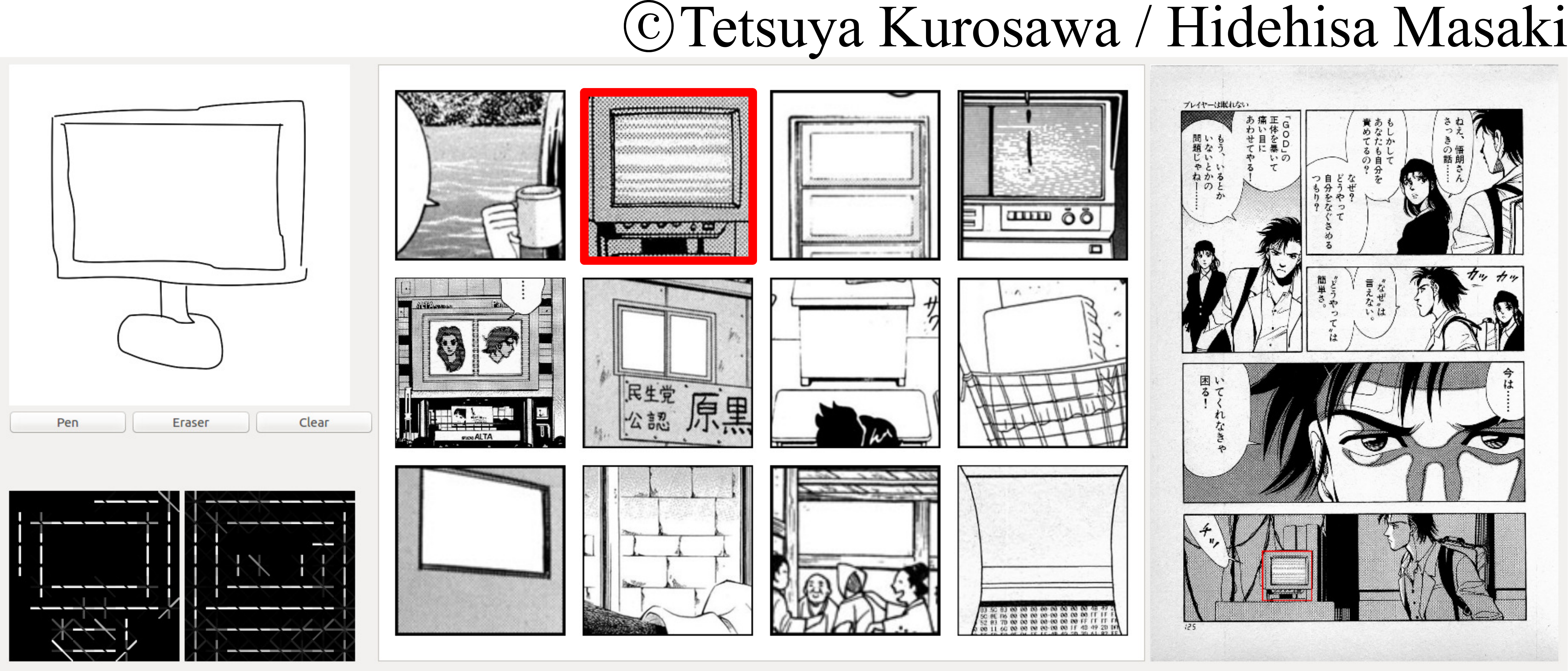}}
	\subfigure[Hourglass]{\label{fig:eitz_hourglass} \includegraphics[width=1.0\linewidth]{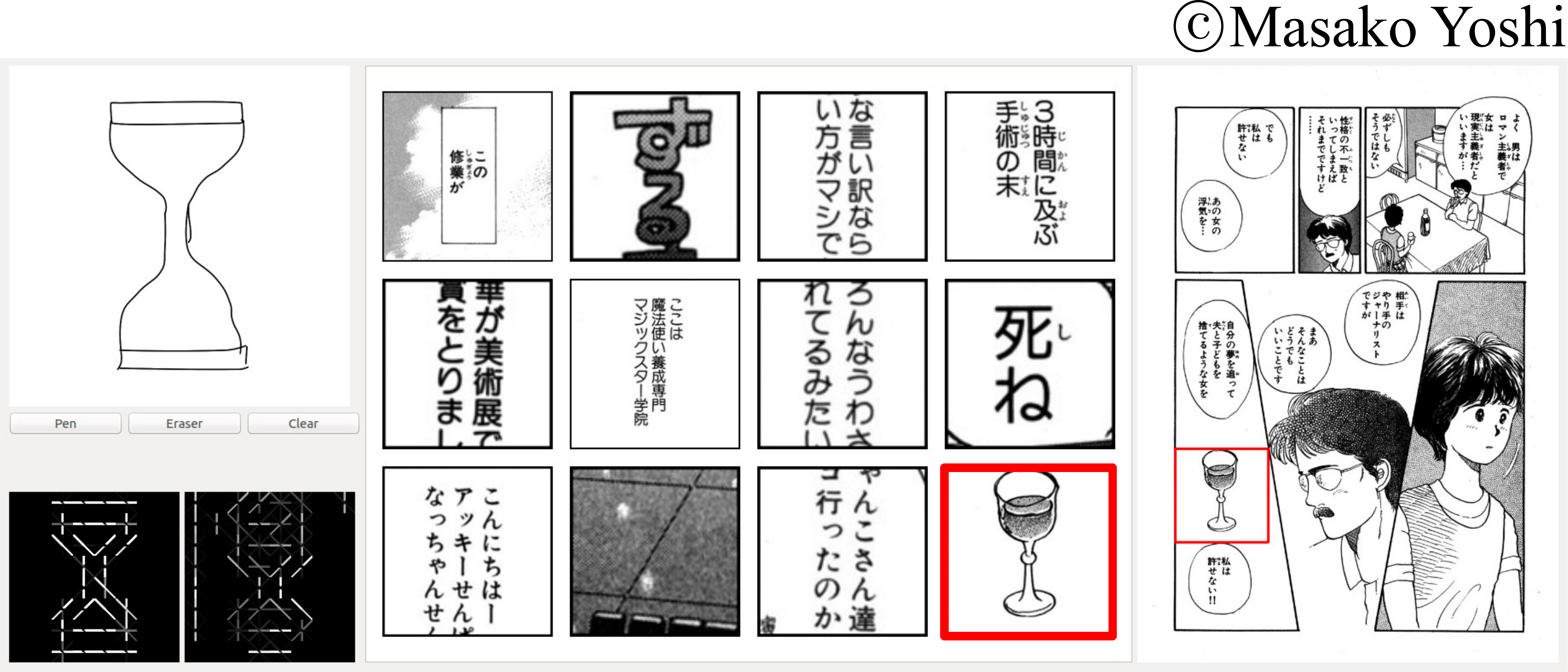}}
	\caption{Results of the subjective study using representative sketches~\cite{eitz2012} as queries.}
	\label{fig:eitz}
\end{figure}

\Fref{fig:eitz_trousers} and \Fref{fig:eitz_computer_monitor} show successful examples.
We could retrieve objects from the Manga109 dataset successfully. 
In particular, the retrieval works well if the target consists of simple geometric shapes such as squares, as shown in \Fref{fig:eitz_computer_monitor}.
This tendency is the same as that for previous sketch-based image retrieval systems~\cite{eitz2011, sun2013a}.
\Fref{fig:eitz_hourglass} shows a failure example, although the retrieved glass is similar to the query.

As can be seen in \Fref{fig:eitz_hourglass}, text regions are sometimes retrieved and placed at the top of the ranking.
Because users usually do not require such results, detecting and eliminating text areas would improve the results, which remains as a future work.

\begin{figure}
	\centering
	\subfigure[]{\label{fig:relevance_feedback_more1} \includegraphics[width=1.0\linewidth]{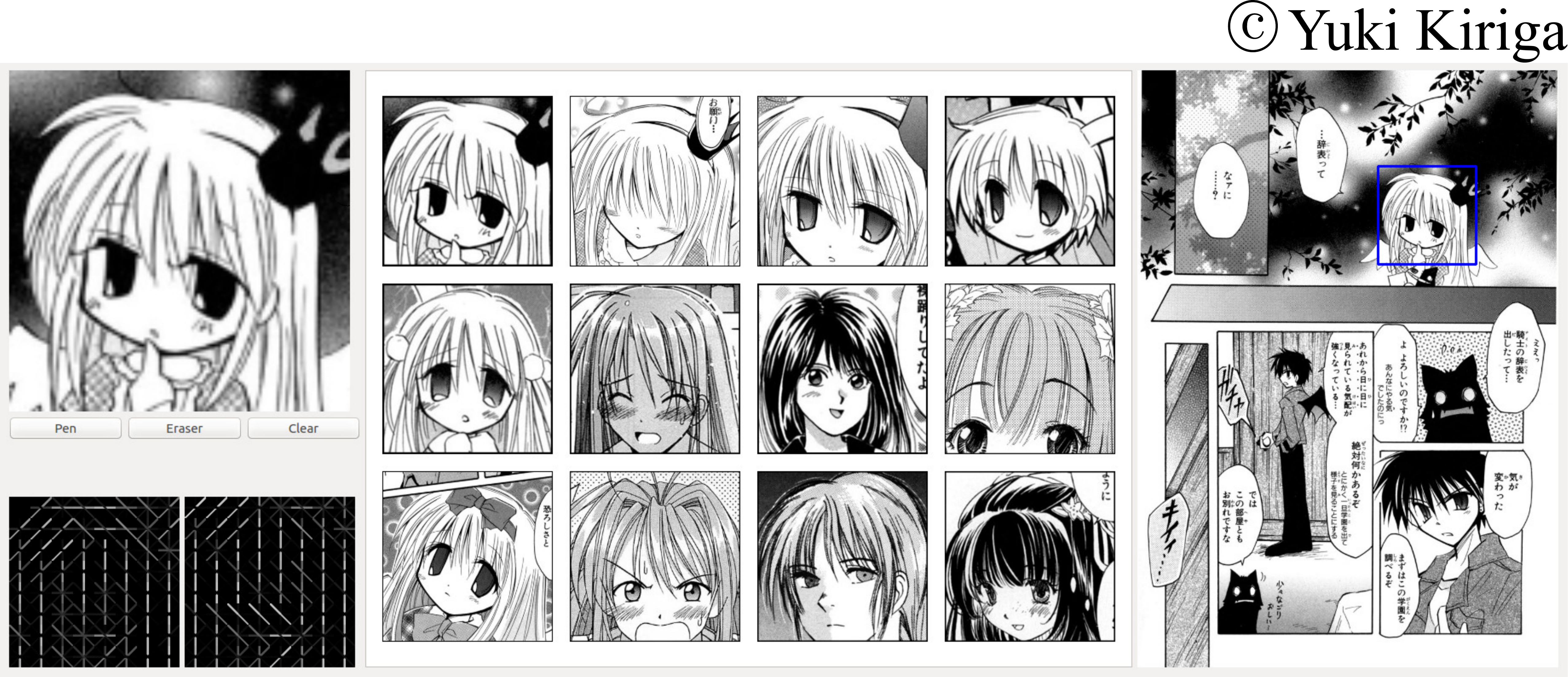}}
	\subfigure[]{\label{fig:relevance_feedback_more2} \includegraphics[width=1.0\linewidth]{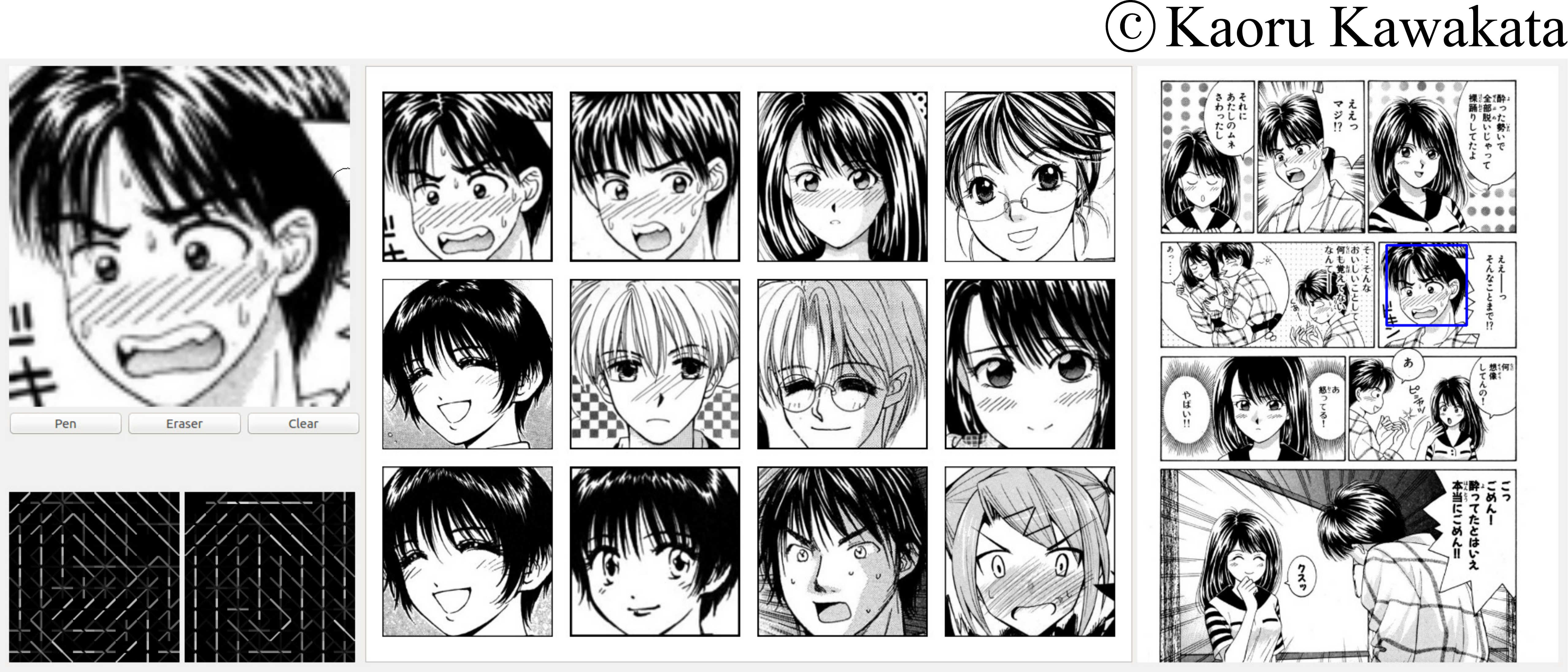}}
	\subfigure[]{\label{fig:relevance_feedback_more3} \includegraphics[width=1.0\linewidth]{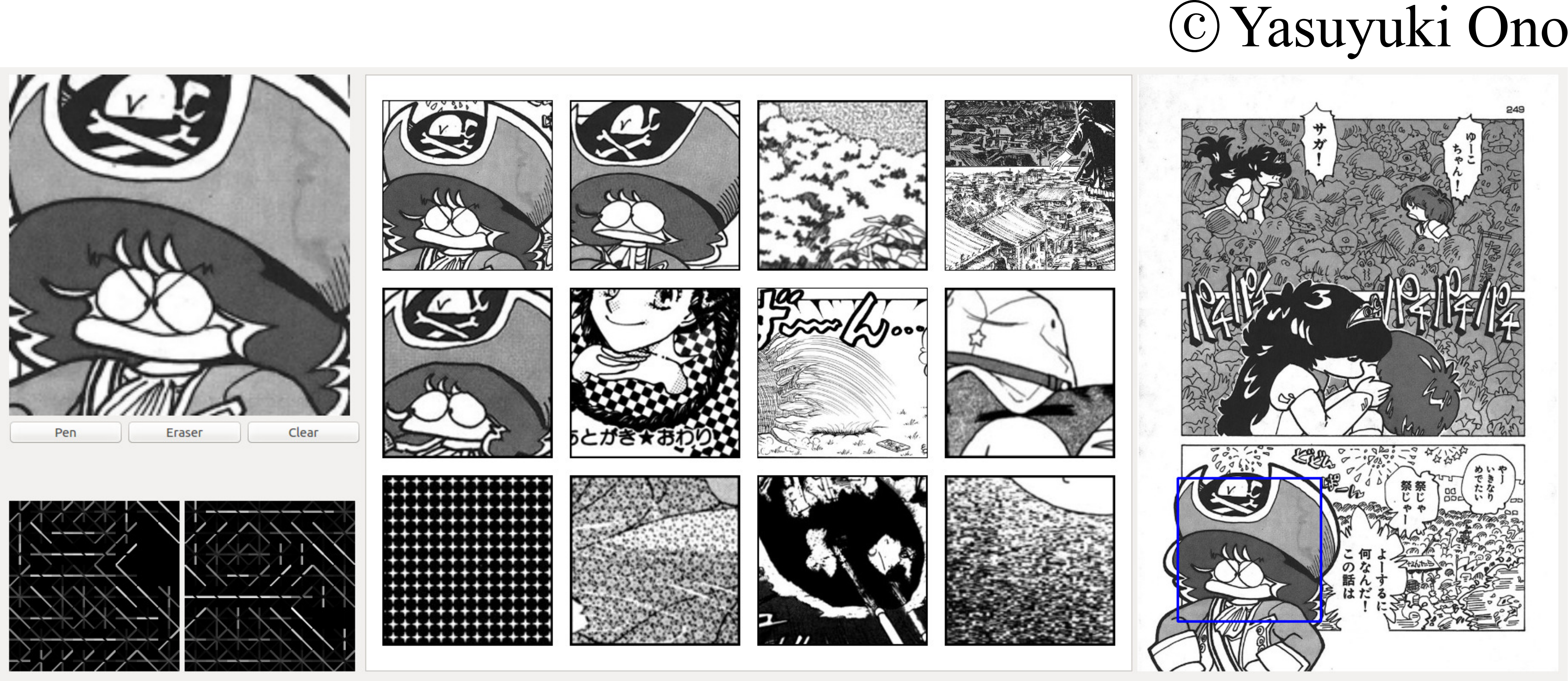}}
	\caption{More results for relevance feedback from Manga109 dataset.}
	\label{fig:relevance_feedback_more}
\end{figure}

\textbf{More results by relevance feedback:}
We show more results from queries of character faces using the proposed relevance feedback in \Fref{fig:relevance_feedback_more}.
We see that the top retrieved results were the same as (or similar to) the characters in the query.
In this case, all the results were drawn by the same author.
Interestingly, our simple edge histogram feature captured the characteristic of authors.
\Fref{fig:relevance_feedback_more2} shows the results of ``blush face.'' In Japanese manga, such blush faces are represented by hatching.
By the relevance feedback, blushed characters were retrieved from various kinds of manga titles.
These character-based retrievals are made possible by content-based search.
This suggests that the proposed query interactions are beneficial for manga search.

\section{Conclusion}
We proposed a sketch-based manga retrieval system and novel query schemes. The retrieval system consists of three steps: labeling of margin areas, EOH feature description, and approximate nearest-neighbor search using product quantization. The query schemes include relevance feedback and query retouch, both of which are new schemes that interactively rerank the retrieved results. We built a new dataset of manga images, Manga109, which consists of 21,142 manga images drawn by professional manga artists. To the best of our knowledge, Manga109 is currently the biggest manga image dataset. It is available to the research community. We conducted a comparative study, localization evaluation, and a large-scale qualitative study. The experiments verified that: (1) the retrieval accuracy of the proposed method is higher than those of existing methods; (2) the proposed method can localize an object instance with reasonable runtime and accuracy; and (3) sketch querying is useful for manga search.

Combining the sketch- and keyword-based searches is a promising direction for future work.

\section*{Acknowledgment}
This work was supported by the Strategic Information and
Communications R\&D Promotion Programme (SCOPE) and JSPS KAKENHI Grant Number 257696.

\ifCLASSOPTIONcaptionsoff
  \newpage
\fi



\bibliographystyle{IEEEtran}
\bibliography{mybibtex}
\end{document}